\documentclass[journal]{IEEEtran}
\ifCLASSINFOpdf
\else
\fi
\usepackage{multirow} 

\usepackage{graphicx} 
\usepackage{amsmath} 
\usepackage{booktabs}
\usepackage{amssymb}

\begin{document}
%
\title{Exploiting Semantic and Pixel Representations for Ultra-Low Bitrate Image Compression}
%
%
%

\author{Hao~Wei,
        Yanhui~Zhou,
        Chenyang Ge,
        Saeed~Anwar,
        and~Ajmal~Mian

\thanks{Hao Wei and Chenyang Ge are with National Key Laboratory of Human-Machine Hybrid Augmented Intelligence, Institute of Artificial Intelligence and Robotics, Xi'an Jiaotong University, Xi'an 710049, China (e-mail: haowei@stu.xjtu.edu.cn; cyge@mail.xjtu.edu.cn).}

\thanks{Yanhui Zhou is with the School of Information and Telecommunication, Xi'an Jiaotong University, Xi'an 710049, China (e-mail: zhouyh@mail.xjtu.edu.cn).}

\thanks{Ajmal Mian and Saeed Anwar are with the Department of Computer Science and Software Engineering, The University of Western Australia, Perth, Crawley, WA 6009, Australia (e-mail:ajmal.mian@uwa.edu.au; saeed.anwar@uwa.edu.au). Ajmal Mian is the recipient of an Australian Research Council Future Fellowship Award (project number FT210100268) funded by the Australian Government.}
}

\maketitle
\begin{abstract}
%
Most existing extreme compression methods fail to achieve an optimal rate–distortion–perception trade-off, as they typically prioritize perceptual fidelity and visual realism over pixel-level accuracy. Consequently, the resulting reconstructions often deviate noticeably from the originals. Ultra-low bitrate image compression is therefore crucial—not only for producing extremely compact representations but also for ensuring that reconstructed images remain semantically coherent and faithful to the source at the pixel level.
To this end, we propose SPRDiff, a diffusion-based compression method that fully leverages both semantic and pixel representations, thereby enhancing reconstruction fidelity under ultra-low bitrate constraints. Specifically, we develop a triple-encoder architecture that utilizes high-fidelity features from the pretrained distortion-oriented and semantic-oriented encoders to compensate for the limited representations extracted by the frozen VAE encoder, thereby improving latent compression and entropy modeling. To further enhance the reconstruction fidelity of diffusion models, we introduce a distortion-aware 
reconstruction module with dual feature extraction. This module not only generates a coarse reconstruction that preserves the main structures, but also provides practical and accurate semantic- and pixel-level conditional signals to guide the diffusion model.
Extensive experiments on benchmark datasets demonstrate that our method outperforms state-of-the-art approaches in the rate-distortion-perception tradeoff at extremely low bitrates (below 0.03 bpp), effectively preserving both perceptual quality and pixel-wise fidelity in the reconstructed images.
We will release the source code and trained models at https://github.com/cshw2021/SPRDiff.

\end{abstract}


\section{Introduction}
\label{introduction}
\IEEEPARstart{C}ompressing images at ultra-low bitrates is of critical importance in environments where storage capacity and communication bandwidth are severely constrained. Such conditions frequently arise in edge computing platforms, remote sensing systems, and large-scale IoT deployments, where devices must transmit visual information under stringent resource limitations. 

\begin{figure*}[!t]
\footnotesize
\centering
    \begin{tabular}{c c c c c c c}
            \multicolumn{3}{c}{\multirow{5}*[45.6pt]{
            \hspace{-2.5mm} \includegraphics[width=0.325\linewidth,height=0.23\linewidth]{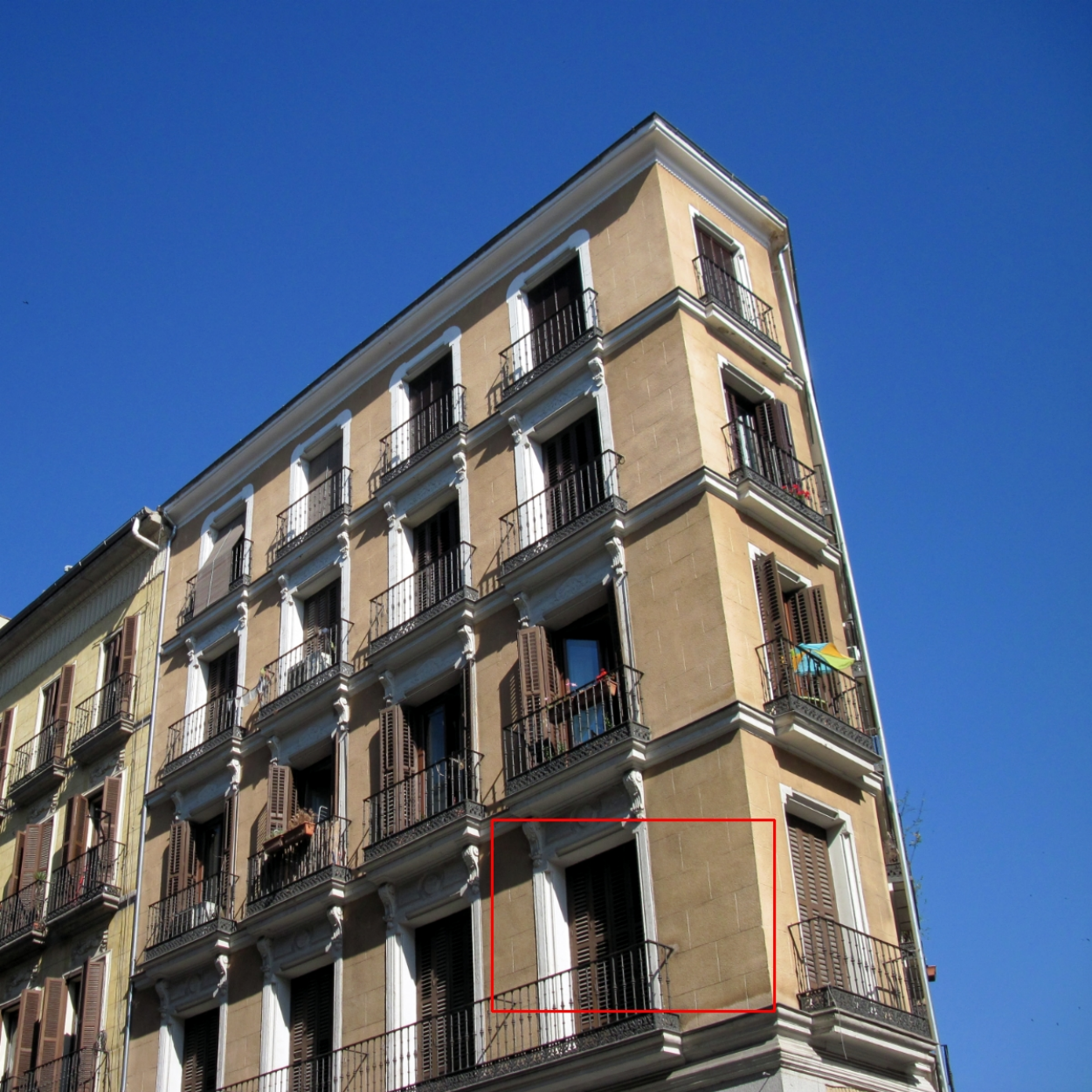}}}
            & \hspace{-4.0mm} \includegraphics[width=0.16\linewidth,height=0.105\linewidth]{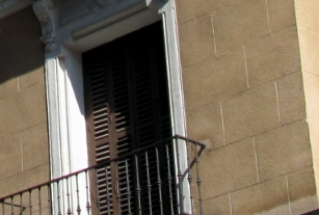}
            & \hspace{-4.0mm} \includegraphics[width=0.16\linewidth,height=0.105\linewidth]{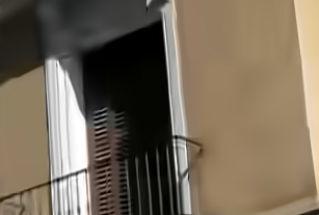}
            & \hspace{-4.0mm} \includegraphics[width=0.16\linewidth,height=0.105\linewidth]{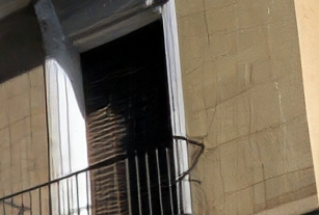}
            & \hspace{-4.0mm} \includegraphics[width=0.16\linewidth,height=0.105\linewidth]{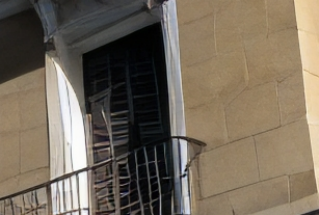}
              \\
    		\multicolumn{3}{c}{~}
            & \hspace{-4.0mm} (a) Original patch (24)
            & \hspace{-4.0mm} (b) VVC (0.0536)
            & \hspace{-4.0mm} (c) MS-ILLM (0.0435)
            & \hspace{-4.0mm} (d) GLC (0.0195) \\		
    	\multicolumn{3}{c}{~}
            & \hspace{-4.0mm} \includegraphics[width=0.16\linewidth,height=0.105\linewidth]{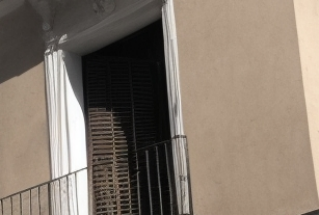}
            & \hspace{-4.0mm} \includegraphics[width=0.16\linewidth,height=0.105\linewidth]{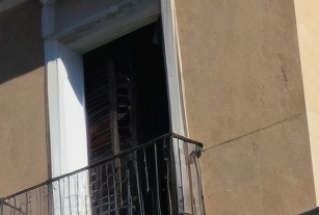}
            & \hspace{-4.0mm} \includegraphics[width=0.16\linewidth,height=0.105\linewidth]{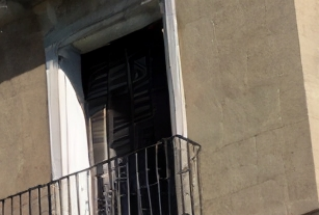}
            & \hspace{-4.0mm} \includegraphics[width=0.16\linewidth,height=0.105\linewidth]{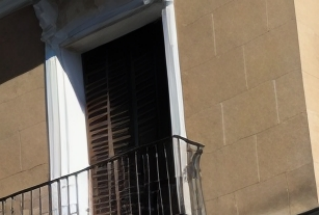}
            \\
    	\multicolumn{3}{c}{\hspace{-4.0mm} Original image}
            & \hspace{-4.0mm} (e) DiffEIC (0.0185)
            & \hspace{-4.0mm} (f) StableCodec (0.0206)
            & \hspace{-4.0mm} (g) OSCAR (0.0313)
            & \hspace{-4.0mm} (h) SPRDiff (\textbf{0.0178})\\

    \end{tabular}
\vspace{-2mm}
\caption{Comparison with state-of-the-art compression methods. The value in parentheses indicates the bpp used by the compressed image. The best value is highlighted in bold. The proposed SPRDiff produces more realistic, high-quality results while preserving intricate structural details at extremely low bitrate e.g. see the perimeter fence and the window blinds.}
\label{vis_com_intro}
\vspace{-3mm}
\end{figure*}
While classical standards such as JPEG~\cite{JPEG} and VVC~\cite{VVC}, as well as learning-based compression methods~\cite{TCM_CVPR2023, FTIC_ICLR2024} optimized for rate-distortion tradeoff, have achieved impressive compression performance, they consistently struggle to maintain visual fidelity at extremely low bitrates (see Fig.~\ref{vis_com_intro}~(b)). In such regimes, the inherent limitations of their representational capacity often result in over-smoothed textures, loss of structural details, and diminished perceptual realism. To address these challenges, recent perceptual image compression approaches based on generative adversarial networks have emerged~\cite{MS-ILLM_ICML2023, GLC_CVPR2024, ICISP_NN2025}. However, they often yield visible artifacts and distorted structures at low-bitrates (see Fig.~\ref{vis_com_intro}~(c)~and~(d)).

Recently, diffusion models have exhibited unparalleled potential for extreme image compression owing to their powerful generative priors~\cite{DiffEIC_TCSVT2024, RDEIC_TCSVT2025, SEDIC_CVPR2025}. For instance, the DiffEIC method~\cite{DiffEIC_TCSVT2024} injects compressed representations into a pretrained stable diffusion model~\cite{SD_CVPR2022} through a lightweight module, formulating the multi-step denoising trajectory as the decoding process. However, in the absence of strong pixel-level constraints, this paradigm is prone to color inconsistencies and the loss of fine structural details (refer to Fig.~\ref{vis_com_intro}~(e)). Additionally, the iterative denoising process involves high latency as well as computational burden, which are both unacceptable for compression tasks. To handle this, several single-step diffusion-based compression approaches have recently been proposed~\cite{StableCodec_ICCV2025, OSCAR_Arxiv2025, SODEC_Arxiv2025}. These methods typically incorporate two types of conditioning signals. On the one hand, they leverage textual semantic cues-such as image-aware prompts~\cite{DiffPC_ICLR2025} or robust, generic prompts (e.g., ``a high-resolution, 8K, ultra-realistic image with sharp focus, vibrant colors, and natural lighting")~\cite{StableCodec_ICCV2025}-to guide the model toward perceptually compelling reconstructions. On the other hand, they introduce visual semantics extracted from degraded observations to better constrain the generative process and enhance reconstruction fidelity~\cite{SODEC_Arxiv2025}. However, due to the lack of effective semantic and pixel-level image conditions, the above-mentioned diffusion-based methods struggle to achieve an optimal rate-distortion-perception trade-off. They either produce overly smoothed outputs or compromise pixel fidelity, leading to reconstructions that drift from the original content, as shown in Fig.~\ref{vis_com_intro}~(f)~and~(g). Therefore, \textit{achieving balanced improvements in both perceptual quality and pixel-level fidelity of reconstructions without favoring one aspect over the other} has emerged as a critical challenge in ultra-low bitrate image compression.

\begin{figure}[htbp]
\centering
\includegraphics[width=0.48\textwidth]{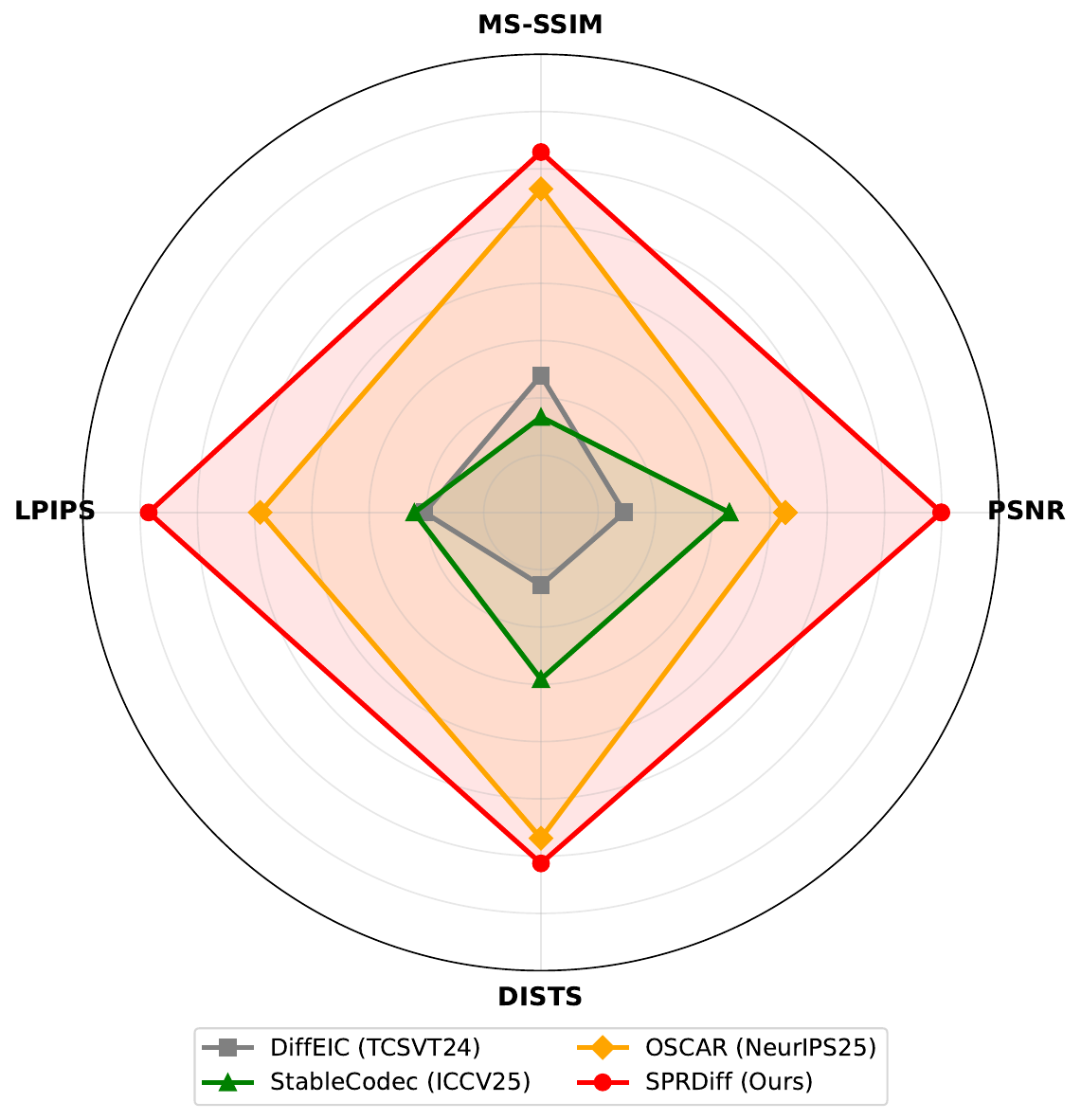}\vspace{-0.4cm}
\caption{Comparison of diffusion-based extreme compression methods on the Kodak dataset. The average bpp values for compression are  0.0313, 0.0201, 0.0186, and \textbf{0.0139}, which correspond to  OSCAR, DiffEIC, StableCodec, and the proposed SPRDiff method, respectively. Our method achieves the best image quality metrics and that too at the highest compression rate.} 
\label{koda_radar}
\end{figure}

To address the aforementioned challenge, we propose \textbf{SPRDiff}, a one-step diffusion codec that fully exploits the semantical and pixel representations for better fidelity reconstruction at both perceptual and pixel levels. First, we establish a triple-encoder architecture by augmenting the Variational Auto Encoder (VAE) with a distortion-oriented compression encoder and a semantic-oriented encoder. The former extracts features instrumental to improving pixel-level fidelity, whereas the latter provides high-level semantic representations. These complementary features are subsequently fused through a lightweight feature fusion block and passed to the latent compression module, which performs transform coding. Second, we develop a distortion-driven reconstruction module that produces a coarse reconstruction with high pixel fidelity. This coarse output is then independently processed by a dedicated lightweight projector and a semantic encoder to produce pixel-level and semantic conditioning signals for the pretrained diffusion model. In doing so, our compression framework fully exploits the generative priors stored in pretrained diffusion models while substantially enhancing both pixel-level fidelity and perceptual reconstruction quality. As illustrated in Fig. \ref{vis_com_intro}(h), the proposed SPRDiff is capable of reconstructing faithful and structurally accurate details even at substantially lower bitrates. Moreover, compared with state-of-the-art diffusion-based approaches, SPRDiff consistently achieves superior perceptual quality and pixel-level fidelity in the extreme bitrate regime, as depicted in Fig. \ref{koda_radar}. 

The contributions of this work are summarized as follows:
\begin{itemize}
    \item We propose a diffusion-based compression framework (SPRDiff) that achieves high pixel-level and perceptual quality even under extremely low bitrates ($< 0.03$ bpp).
    \item We develop a triple-encoder structure that fully explores complementary representations: a distortion-oriented encoder for high-fidelity features and a pretrained semantic-oriented encoder for rich semantic priors.
    \item We design a distortion-aware reconstruction module together with dual-feature extraction, enabling the generation of effective semantic- and pixel-level conditions and substantially enhancing diffusion-based reconstruction fidelity.   
\end{itemize}



\section{Related Works}
\label{related_works}
\begin{figure*}[t]
\centering
\includegraphics[width=1.0\textwidth]{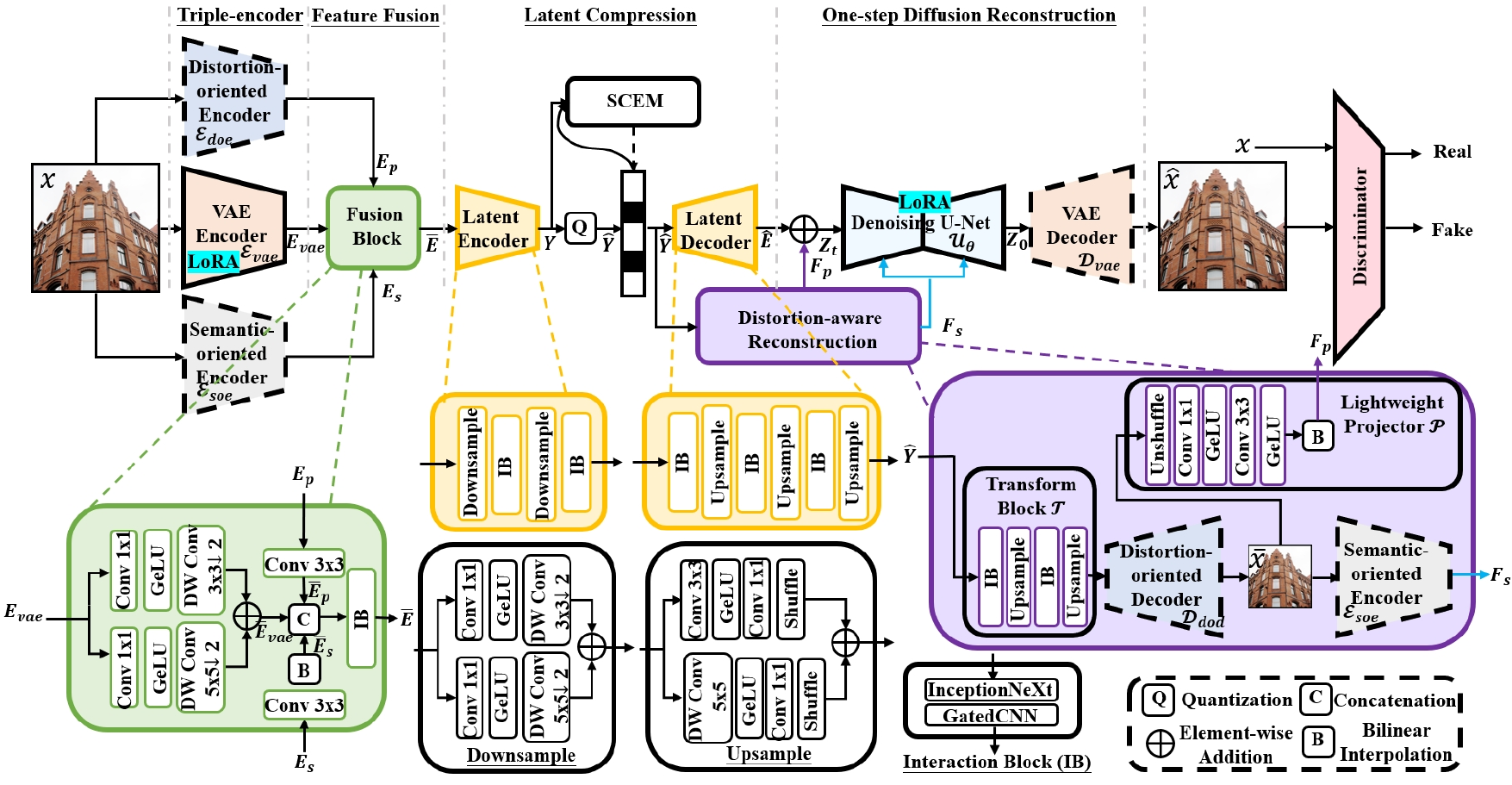}
\caption{The overall framework of the proposed SPRDiff consists of a triple-encoder, a feature fusion block, a latent compression module, and a one-step diffusion reconstruction module. The triple-encoder extracts complementary feature representations $\{E_p, E_{vae}, E_s\}$, which are subsequently fused and compressed. During decoding, the one-step diffusion module reconstructs the final image $\hat{x}$, conditioned on the outputs $\{F_p, F_s\}$ from the distortion-aware reconstruction module. SCEM denotes the space-channel entropy model adapted from~\cite{ELIC_CVPR2022}. Moreover, the discriminator provides adversarial supervision, encouraging perceptually realistic reconstructions. Note that the modules outlined with dashed borders remain frozen during training.} 
\label{arch}
\end{figure*}
\subsection{Learned Image Compression Methods}
Learned image compression aims to achieve a significant rate-distortion tradeoff according to Shannon's rate-distortion theory~\cite{SIT_1999}. In~\cite{balle2017end}, Ball{\'e}~et~al. propose the first end-to-end neural network-based compression model, which mainly contains analysis and synthesis transform modules. The former converts images into a compact representation for entropy modeling, and the latter transforms latent features into original images. Subsequently, some works began to focus on improving nonlinear transforms~\cite{cheng2020learned, xie2021enhanced, zou2022devil, LLIC_TMM2024}. For example, a mixed CNN-Transformer architecture is proposed in ~\cite{TCM_CVPR2023} to enable local and non-local modeling. FTIC method~\cite{FTIC_ICLR2024} decomposes features with different frequency components by employing window attention with various window sizes. To mitigate the quadratic complexity of transformers and maintain the non-local modeling ability, MambaIC~\cite{MambaIC_CVPR2025} integrates State Space Models (SSMs) into nonlinear transforms, while LALIC~\cite{LALIC_CVPR2025} leverages the Receptance Weighted Key Value (RWKV) model for capturing long-range dependencies. Apart from improving the transform modules, much research has focused on reducing the bitstream using advanced entropy modeling~\cite{minnen2018joint, guo2021causal, he2021checkerboard, minnen2020channel}. Jiang~et~al.~\cite{MLIC_ACMMM2023, MLIC++_2025} use the multi-reference entropy modeling to capture the local, global, and channel-wise correlations within entropy model. He~et~al.~\cite{ELIC_CVPR2022} reduce the bitrate by combining the checkboard partition and channel-wise context modeling. In~\cite{DCAE_CVPR2025}, Lu~et~al. adopt dictionary learning to establish external dependencies between the latent representation and prior information from external training data. However, the above-mentioned distortion-oriented methods struggle to compress images at extremely low bitrates, resulting in overly smoothed reconstruction. 
\subsection{Conventional Perceptual Image Compression}
Perceptual image compression~\cite{blau2019rethinking} focuses on achieving a tradeoff among rate, distortion, and perception, while alleviating the problem of excessive image smoothing and low statistical fidelity of reconstructions at low bitrates. Mentzer~et~al.~\cite{HiFiC_NeurIPS2020} introduce a generative adversarial network (GAN) with a perceptual loss to improve visual fidelity. Subsequently, some works explore the variants of discriminators, such as local label prediction~\cite{MS-ILLM_ICML2023}, realism guidance~\cite{MRIC_CVPR2023}, and implicit priors guidance~\cite{ICISP_NN2025}. In~\cite{TACO_ICML2024}, Lee et. al. improve both pixel-level and perceptual fidelity using the text to guide the encoding process. Inspired by the success of diffusion models in the generation task, recent work has attempted to apply them to generative image compression~\cite{DIRAC_Arxiv2023, CGDM_ACMMM2024, Zhou_AAAI2025}. In~\cite{CDC_NeurIPS2023}, Yang~et~al. employ the diffusion model as the decoder of the autoencoder-based compression. Ma~et~al.~\cite{CorrDiff_ICML2024} correct the sampling process of the diffusion model using a privileged end-to-end decoder, further enhancing the reconstruction quality. These approaches, targeting conventional bitrates (over 0.1 bpp), however, commonly suffer from degradations that manifest as blurring, texture loss, and color shifts when applied to extremely low bitrates.

\subsection{Extremely-low Bitrate Image Compression}
The majority of extremely-low bitrate image compression methods often leverage GAN or diffusion models to achieve visually pleasing reconstruction~\cite{Jiang_AAAI2023, PerCo_ICLR2023, Text_Sketch, MISC_TIP24, CG_TIP24}. In~\cite{Mao_DCC2024}, Mao~et~al. propose adjusting the codebook size using K-means clustering in the pretrained VQGAN~\cite{VQGAN_CVPR2021, EIR_TCSVT2024}. Jia~et~al.~\cite{GLC_CVPR2024} perform latent transform in the latent space of VQGAN, achieving significant compression performance. To complement the details lost in the pretrained VQGAN, Hybridflow~\cite{Hybridflow_ACMMM2024} and DLF~\cite{DLF_Arxiv2025} respectively adopt continuous features from the MLIC encoder~\cite{MLIC_ACMMM2023} and the dynamic detail encoder. Both DiffEIC~\cite{DiffEIC_TCSVT2024} and RDEIC~\cite{RDEIC_TCSVT2025} adopt compressed content to guide a frozen diffusion model through a lightweight control module. Song~et~al.~\cite{SEDIC_CVPR2025} employ large multimodal models~\cite{Murai_VCIP2024} to disentangle the image into different compact semantic representations, which are then compressed separately. However, these approaches fail to reconstruct images with high pixel-level fidelity due to insufficient constraints in the pixel space. Additionally, the multi-step denoising process is computationally expensive, resulting in inefficient decoding~\cite{RDULIC_ICML2025}. To improve the efficiency and practicality of diffusion-based extreme compression methods, some studies have begun to improve the sampling efficiency with only one step~\cite{DiffO_Arxiv2025, SODEC_Arxiv2025}. In~\cite{StableCodec_ICCV2025}, Zhang~et~al. use the generative priors from pretrained SD-Turbo~\cite{SD-Turbo_ECCV2024} that yield consistent perceptual reconstructions in a single denoising step. Guo~et~al.~\cite{OSCAR_Arxiv2025} achieve a multi-rate diffusion-based codec by bridging quantization and forward diffusion. However, current diffusion-based extreme image compression algorithms struggle to achieve an optimal balance between distortion and perceptual quality when compressing images at ultra-low bitrates, often sacrificing pixel fidelity in pursuit of perceptual quality in the reconstruction. Improving both perceptual and pixel-level fidelity is therefore essential for diffusion-based ultra-low bitrate compression.

\section{Methodology}
\label{methodology}
Our goal is to develop an effective extreme image compression method capable of operating at 0.03 bpp or lower while preserving perceptual quality and pixel-level fidelity in the reconstructed images. The overall framework of the proposed model is illustrated in Fig.~\ref{arch}. In the following, we describe each component of the framework in detail.

\subsection{Complementary Feature Extraction and Fusion}
\subsubsection{Triple-encoder}
\begin{figure}[h]
\footnotesize
\centering
    \begin{tabular}{c c c}
            \multicolumn{1}{c}{\multirow{1}*[61pt]{
            \hspace{-4.0mm} \includegraphics[width=0.44\linewidth,height=0.58\linewidth]{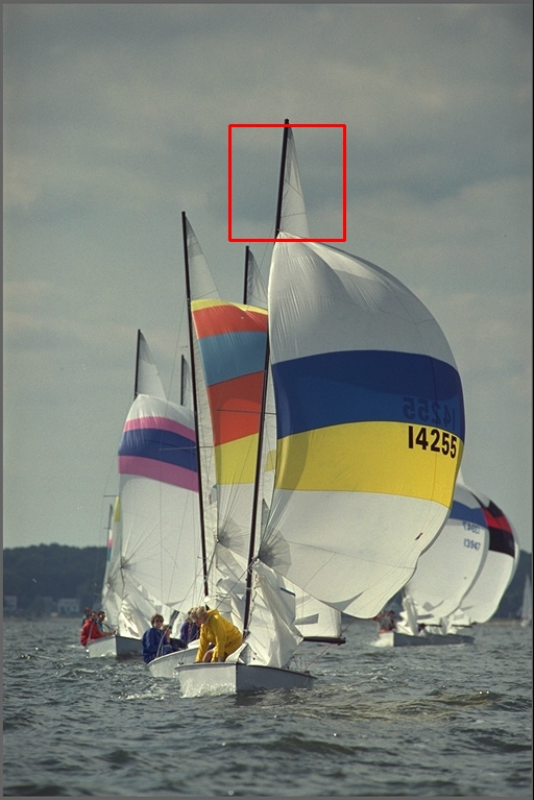}}}
            & \hspace{-4.0mm} \includegraphics[width=0.27\linewidth,height=0.27\linewidth]{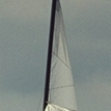}
            & \hspace{-4.0mm} \includegraphics[width=0.27\linewidth,height=0.27\linewidth]{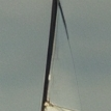}\\
    		\multicolumn{1}{c}{~}
            & \hspace{-4.0mm} (a) Original patch
            & \hspace{-4.0mm} (b) w/ only $\mathcal{E}_{vae}$ \\
    	\multicolumn{1}{c}{~}
            & \hspace{-4.0mm} \includegraphics[width=0.27\linewidth,height=0.27\linewidth]{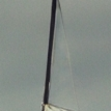}
            & \hspace{-4.0mm} \includegraphics[width=0.27\linewidth,height=0.27\linewidth]{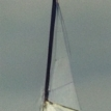} \\
    	\multicolumn{1}{c}{\hspace{-4.0mm} Original image}
            & \hspace{-4.0mm} (c) w/ $\mathcal{E}_{vae}$ \& $\mathcal{E}_{doe}$
            & \hspace{-4.0mm} (d) w/ triple-encoder\\
    \end{tabular}
\vspace{-2mm}
\caption{Visual comparisons of reconstructed images at around 0.02 bpp using different encoders.}
\label{vae_com}
\vspace{-3mm}
\end{figure}

Recent methods commonly rely on a single frozen VAE encoder to extract latent representations, which inevitably leads to substantial spatial information loss~\cite{RDEIC_TCSVT2025, RDULIC_ICML2025}. Furthermore, because this encoder is not explicitly tailored for compression, its latent space is inherently misaligned with the entropy models used in learned image compression. To overcome these limitations, we propose a triple-encoder architecture that integrates a distortion-oriented encoder $\mathcal{E}_{doe}$ and a semantic-oriented encoder $\mathcal{E}_{soe}$ alongside the pretrained VAE $\mathcal{E}_{vae}$. This design enriches and refines the latent representation, ensuring improved fidelity and compatibility with subsequent compression. Specifically, given an input image $x$, the triple-encoder produces three complementary feature representations:
\begin{equation}
    E_p = \mathcal{E}_{doe}(x), \quad E_{vae} = \mathcal{E}_{vae}(x), \quad E_{s} = \mathcal{E}_{soe}(x),
\end{equation}
where $\{E_p, E_{vae}, E_s\}$ denote the distortion-oriented, VAE-based, and semantic-oriented features, respectively. As illustrated in Fig.~\ref{vae_com}, the proposed triple-encoder yields a more faithful structural reconstruction compared to using the VAE encoder alone.

\subsubsection{Lightweight Feature Fusion Module}
Since the output features of the proposed triple-encoder differ in spatial resolution, they cannot be directly fused in their original form. To alleviate this, we design an effective feature fusion module, as illustrated in Fig.~\ref{arch}. 
First, we align the spatial dimensions of the triple-encoder outputs as follows:
\begin{equation}
\begin{split}
    \bar{E}_{vae} = \mathcal{N}_3(E_{vae}) + \mathcal{N}_5(E_{vae}), \\
    \bar{E}_{p} = \mathbf{W}_{2d}(E_p), \quad \bar{E}_s = \textbf{B}(\mathbf{W}_{2d}(E_s)),
\end{split}
\end{equation}
where $\mathcal{N}_i(\cdot)$ denotes a plain network composed of a 1$\times$1 convolution followed by an $i\times i$ depth-wise convolution with stride 2, with a GeLU activation applied in between. $\mathbf{W}_{2d}(\cdot)$ represents a standard 3$\times$3 convolution layer, and $\textbf{B}(\cdot)$ denotes bilinear interpolation. After obtaining the refined intermediate features $\{\bar{E}_{vae}, \bar{E}_p, \bar{E}_s\}$, we concatenate them and feed the result into a basic interaction block $\mathcal{M}$:
\begin{equation}
\label{FB}
    \bar{E} = \mathcal{M}(\textbf{C}(\bar{E}_{vae}, \bar{E}_p, \bar{E}_s)),
\end{equation}
where \textbf{C} denotes the concatenation operator. The module 
$\mathcal{M}(\cdot)$ is constructed by cascading InceptionNeXt~\cite{Inceptionnext} and GatedCNN~\cite{MambaOut} blocks, enabling more effective feature interaction and integration. Lastly, $\bar{E}$ is the fused output feature.

\subsection{Latent Compression}
To obtain a more compact representation $Y$, we compress the refined feature $\bar{E}$ derived from Eq.~(\ref{FB}) using a latent encoder $\mathcal{E}_l$. The resulting latent $Y$ is then quantized into $\hat{Y}$ for entropy coding. Subsequently, a latent decoder $\mathcal{D}_l$ reconstructs the latent feature $\hat{E}$ from $\hat{Y}$. The overall process can be expressed as:

\begin{equation}
\label{lc}
\begin{split}
     Y &= \mathcal{E}_l(\bar{E}),\\  
     \hat{Y}&=\textbf{Q}(Y),\\  \hat{E}&=\mathcal{D}_l(\hat{Y}), 
\end{split}
\end{equation}
where $\textbf{Q}$ denotes the quantization operation.

\subsection{Distortion-aware Reconstruction with Dual Extraction}
\begin{figure}[!t]
\footnotesize
\centering
    \begin{tabular}{ccc}
            \includegraphics[width=0.317\linewidth,height=0.2\linewidth]{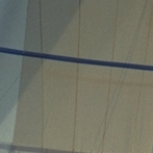}
            \includegraphics[width=0.317\linewidth,height=0.2\linewidth]{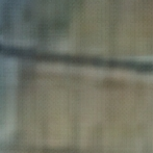}
            \includegraphics[width=0.317\linewidth,height=0.2\linewidth]{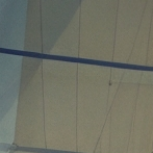} \\
            \makebox[0.317\linewidth]{(a) Original}
            \makebox[0.317\linewidth]{(b) Coarse result $\bar{x}$}
            \makebox[0.317\linewidth]{(c) Fine result $\hat{x}$}	 \\
    \end{tabular}
\vspace{-2mm}
\caption{Reconstruction results at approximately 0.03 bpp. Panel (b) preserves the primary structural content, while panel (c) further provides more accurate fine-grained details.}
\label{dor_com}
\vspace{-3mm}
\end{figure}
Current approaches typically use the decompressed latent $\hat{E}$ directly as a condition to guide the diffusion-based reconstruction, often through a lightweight control module~\cite{DiffEIC_TCSVT2024, RDEIC_TCSVT2025}. However, due to the absence of effective supervisory constraints, the decompressed latent becomes increasingly diluted and ambiguous under extreme compression, failing to preserve sufficient information to characterize the original image uniquely. This inevitably leads to degraded reconstruction fidelity. Inspired by neural compression methods~\cite{MLIC++_2025} optimized for the rate-distortion tradeoff, which, despite losing fine textures, still preserve the main structural content at extremely low bitrates, we develop a distortion-aware reconstruction module. This module enhances retention of both semantic cues and pixel-level structural information, thereby providing more effective conditioning for subsequent diffusion-based reconstruction. The architecture is shown in Fig.~\ref{arch}.

First, we apply the transform block $\mathcal{T}$ to the quantized latent $\hat{Y}$ provided by Eq.~(\ref{lc}). Then the resultant features are sent to the pretrained distortion-oriented decoder $\mathcal{D}_{dod}$ to get the coarse reconstruction $\hat{x}$, which contains the main structural information of the original input (see Fig.~\ref{dor_com}(b)). Finally, we extract the pixel-level condition $F_p$ and semantic condition $F_s$ through a lightweight projector $\mathcal{P}$ and a pretrained semantic-oriented encoder $\mathcal{E}_{soe}$, respectively. The whole process can be formulated as:
\begin{equation}
\label{dor}
\begin{split}
        \bar{x}&=\mathcal{D}_{dod}(\mathcal{T}(\hat{Y})), \\F_p&=\mathcal{P}(\bar{x}), \quad F_s=\mathcal{E}_{soe}(\bar{x}).
\end{split}
\end{equation}

\subsection{One-step Diffusion Reconstruction}
Given the decompressed latent $\hat{E}$ and the pixel-wise condition $F_p$, we initially combine them through element-wise addition to derive a refined latent $Z_t$, which functions as an analogue to a noisy latent at the timestep $t$. During the subsequent diffusion procedure, the semantic features $F_s$ are incorporated as additional conditioning to aid in extracting informative and semantically relevant cues from the diffusion model's latent representation. Notably, cross-attention layers facilitate effective interaction between $F_s$ and the latent features of the diffusion model, thereby enabling the model to capture and integrate semantically pertinent information. Within this diffusion framework, the accurately reconstructed latent representation $Z_0$ with high fidelity is obtained through a single denoising step, as described below.   

\begin{equation}
    Z_0 = \frac{1}{\sqrt{\bar{\alpha_t}}}Z_t - \frac{\sqrt{1-\bar{\alpha_t}}}{\sqrt{\bar{\alpha_t}}}\mathcal{U}_\theta(Z_t, F_s, t),
\end{equation}
where $\mathcal{U}_\theta$ denotes the denoising U-Net, and $\bar{\alpha}_t$ is the noise scheduler of the forward diffusion process of the diffusion model. Finally, the reconstructed image $\hat{x}$ is obtained using the frozen VAE decoder $\mathcal{D}_{vae}$, i.e., $\hat{x}=\mathcal{D}_{vae}(Z_0)$. As shown in Fig.~\ref{dor_com}(c), the diffusion model can generate structural details that closely match those of the original image at extremely low bitrates.

\subsection{Loss Functions}
To train the proposed compression framework, we adopt the following rate–distortion–perception objective:
\begin{equation}
\label{stage1}
\begin{split}
    L_{stage1} &=\lambda_d \|\hat{x}-x\|_2^2 +\lambda_l \|\phi(\hat{x})-\phi(x)\|_2^2\\&+\lambda_i\|\bar{x}-x\|_2^2+\lambda_bR(\hat{Y}),  
\end{split}
\end{equation}
where $\|\cdot\|_2^2$ is the mean squared error, $\phi(\cdot)$ is a pretrained VGG network used for perceptual feature extraction, and $R(\cdot)$ represents the estimated bitrate. The values of the hyperparameters $\{\lambda_d, \lambda_l, \lambda_i, \lambda_b\}$ are provided in Section~\ref{implemental_details}. In addition, we employ an adversarial distillation strategy by performing adversarial training to reduce the distributional gap between reconstructed and original images. The adversarial loss is defined as:
\begin{equation}
    L_{adv} = -\mathbb{E}(log\mathcal{H}(\hat{x})),
\end{equation}
where $\mathcal{H}$ denotes the discriminator. Hence, the full optimization objective for the compression framework in the second stage is given by:
\begin{equation}
\label{stage2}
    L_{stage2} = L_{stage1}+\lambda_aL_{adv},
\end{equation}
where $\lambda_a$ is a weighting hyperparameter.
To constrain the discriminator, we apply the loss as:
\begin{equation}
    L_{dis} = \mathbb{E}[log\mathcal{H}(x)] + \mathbb{E}[log(1-\mathcal{H}(\hat{x}))].
\end{equation}
\section{Experiments}
\label{experiments}
\subsection{Experimental Settings}
\begin{figure*}[htbp]
\centering
\includegraphics[width=1.0\textwidth]{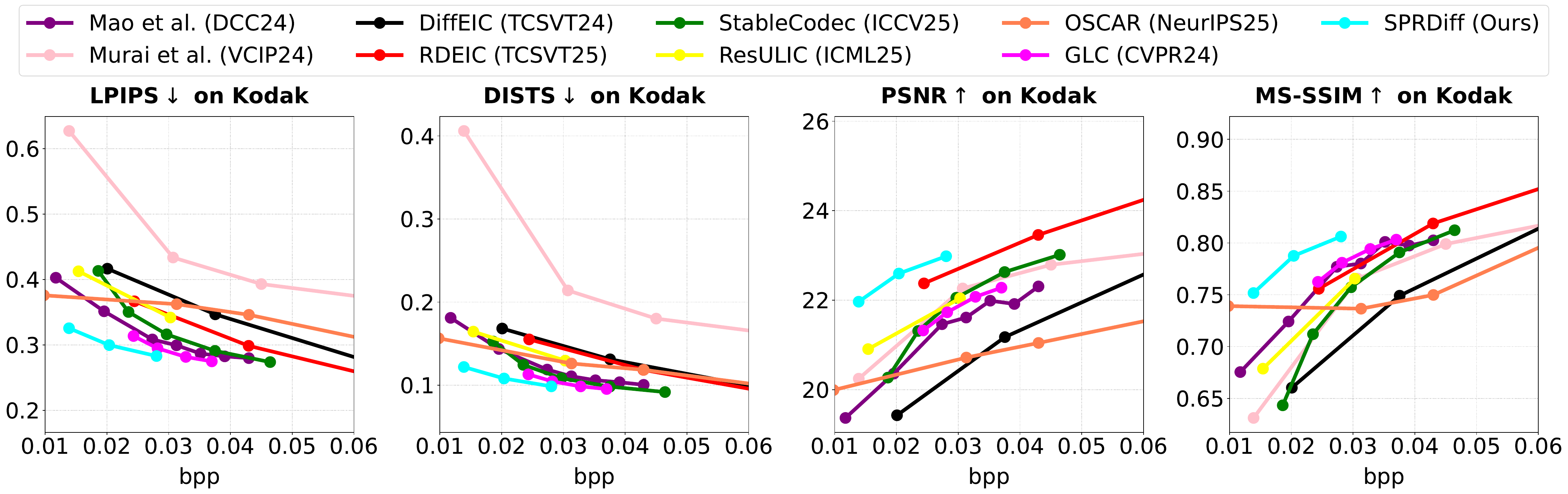} 
\includegraphics[width=1.0\textwidth]{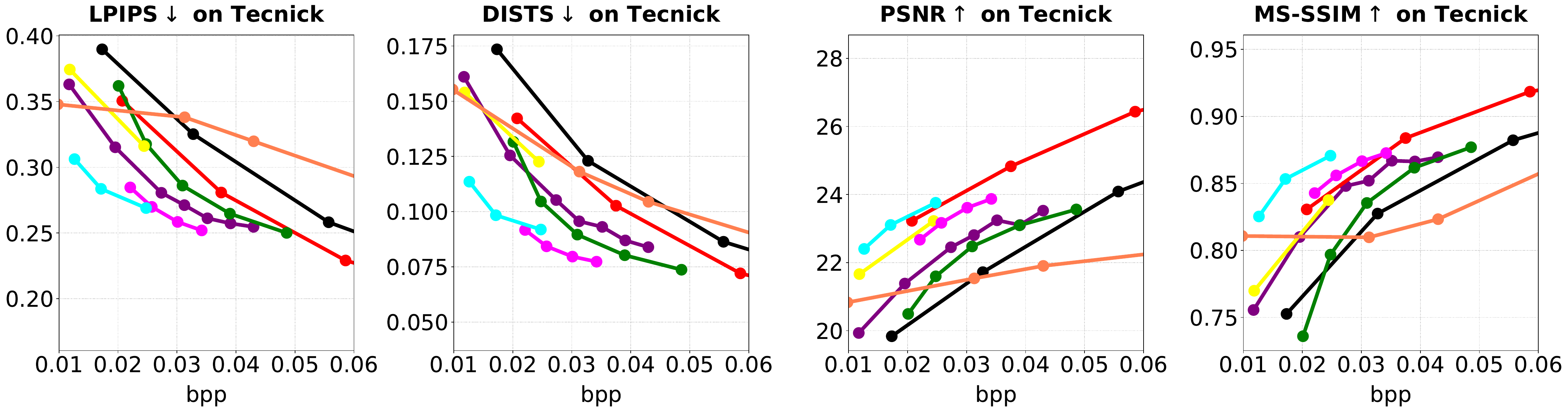} 
\includegraphics[width=1.0\textwidth]{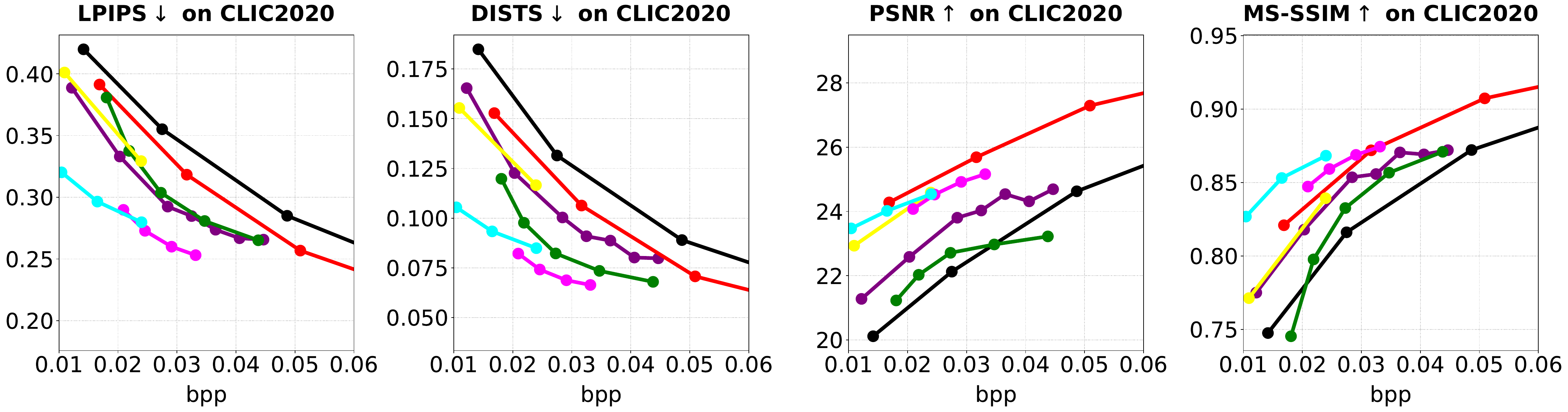} 
\includegraphics[width=1.0\textwidth]{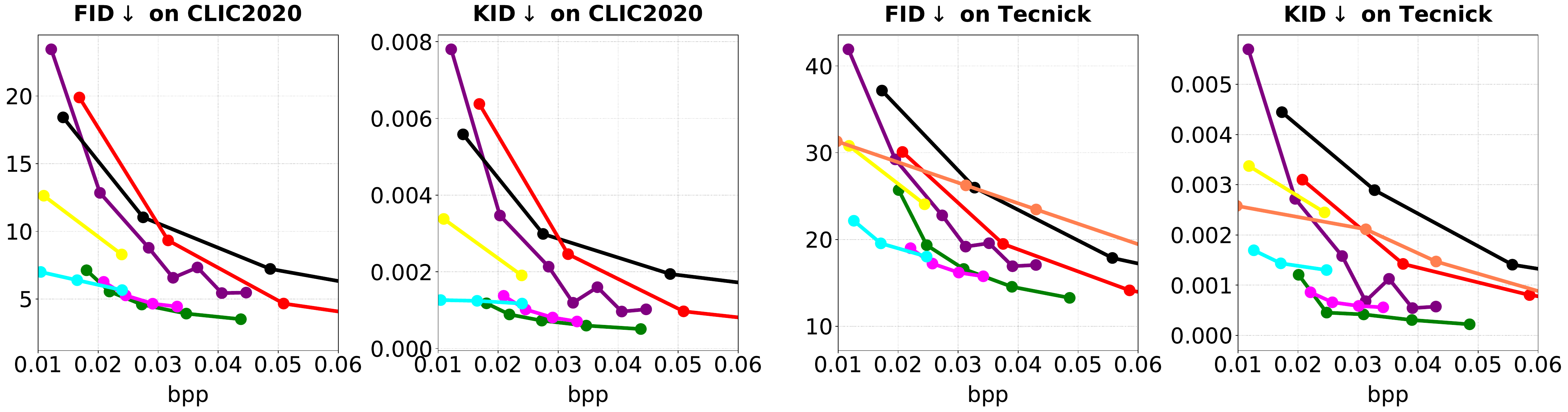} \vspace{-0.5cm}
\caption{Comparison of extreme image compression methods on benchmark datasets. Lower LPIPS, DISTS, FID, and KID values, and higher PSNR and MS-SSIM values, are desirable, along with a lower bpp.} 
\label{quan_com}   
\end{figure*}
\begin{figure*}[!t]
\footnotesize
\centering
    \begin{tabular}{c c c c c c c}
            \multicolumn{3}{c}{\multirow{5}*[45.6pt]{
            \hspace{-2.5mm} \includegraphics[width=0.32\linewidth,height=0.23\linewidth]{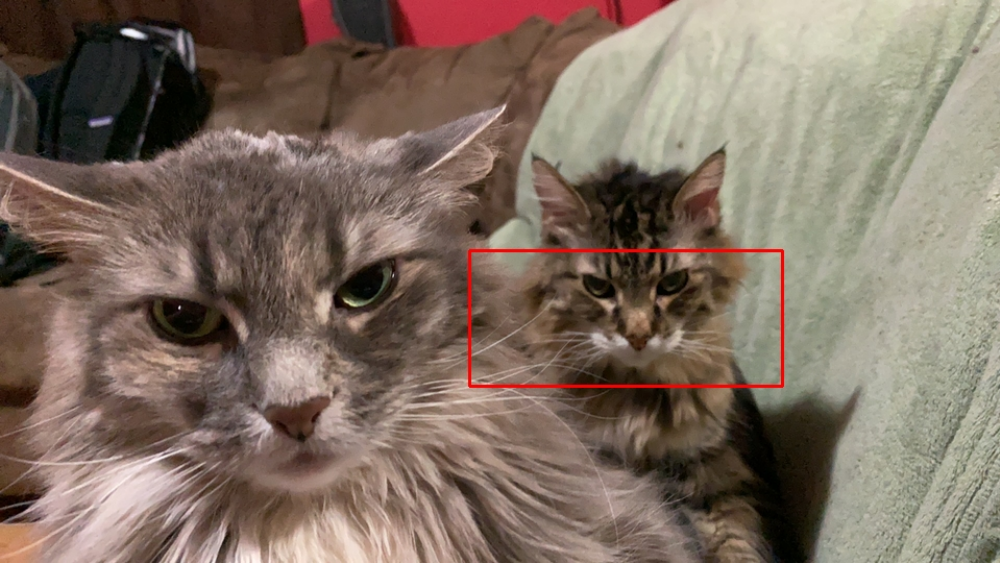}}}
            & \hspace{-4.0mm} \includegraphics[width=0.16\linewidth,height=0.105\linewidth]{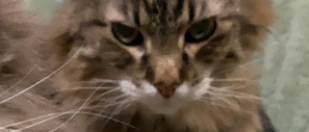}
            & \hspace{-4.0mm} \includegraphics[width=0.16\linewidth,height=0.105\linewidth]{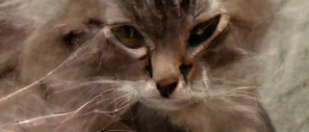}
            & \hspace{-4.0mm} \includegraphics[width=0.16\linewidth,height=0.105\linewidth]{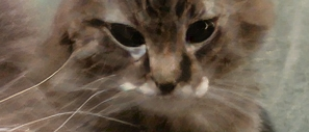}
            & \hspace{-4.0mm} \includegraphics[width=0.16\linewidth,height=0.105\linewidth]{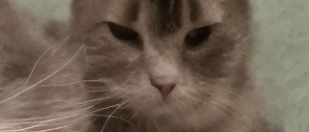}
              \\
    		\multicolumn{3}{c}{~}
            & \hspace{-4.0mm} (a) Original patch (24)
            & \hspace{-4.0mm} (b) Mao et al. (0.0247)
            & \hspace{-4.0mm} (c) GLC (0.0279)
            & \hspace{-4.0mm} (d) DiffEIC (0.0166) \\		
    	\multicolumn{3}{c}{~}
            & \hspace{-4.0mm} \includegraphics[width=0.16\linewidth,height=0.105\linewidth]{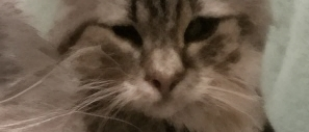}
            & \hspace{-4.0mm} \includegraphics[width=0.16\linewidth,height=0.105\linewidth]{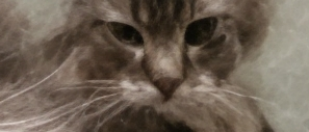}
            & \hspace{-4.0mm} \includegraphics[width=0.16\linewidth,height=0.105\linewidth]{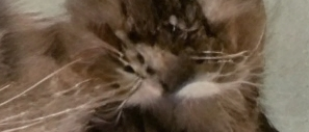}
            & \hspace{-4.0mm} \includegraphics[width=0.16\linewidth,height=0.105\linewidth]{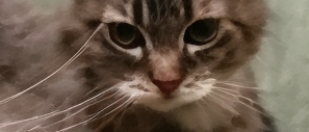}
            \\
    	\multicolumn{3}{c}{\hspace{-4.0mm} Original image}
            & \hspace{-4.0mm} (e) ResULIC (0.0253)
            & \hspace{-4.0mm} (f) RDEIC (0.0198)
            & \hspace{-4.0mm} (g) StableCodec (0.0267)
            & \hspace{-4.0mm} (h) SPRDiff (\textbf{0.0157})\\

            \multicolumn{3}{c}{\multirow{5}*[45.6pt]{
            \hspace{-2.5mm} \includegraphics[width=0.325\linewidth,height=0.23\linewidth]{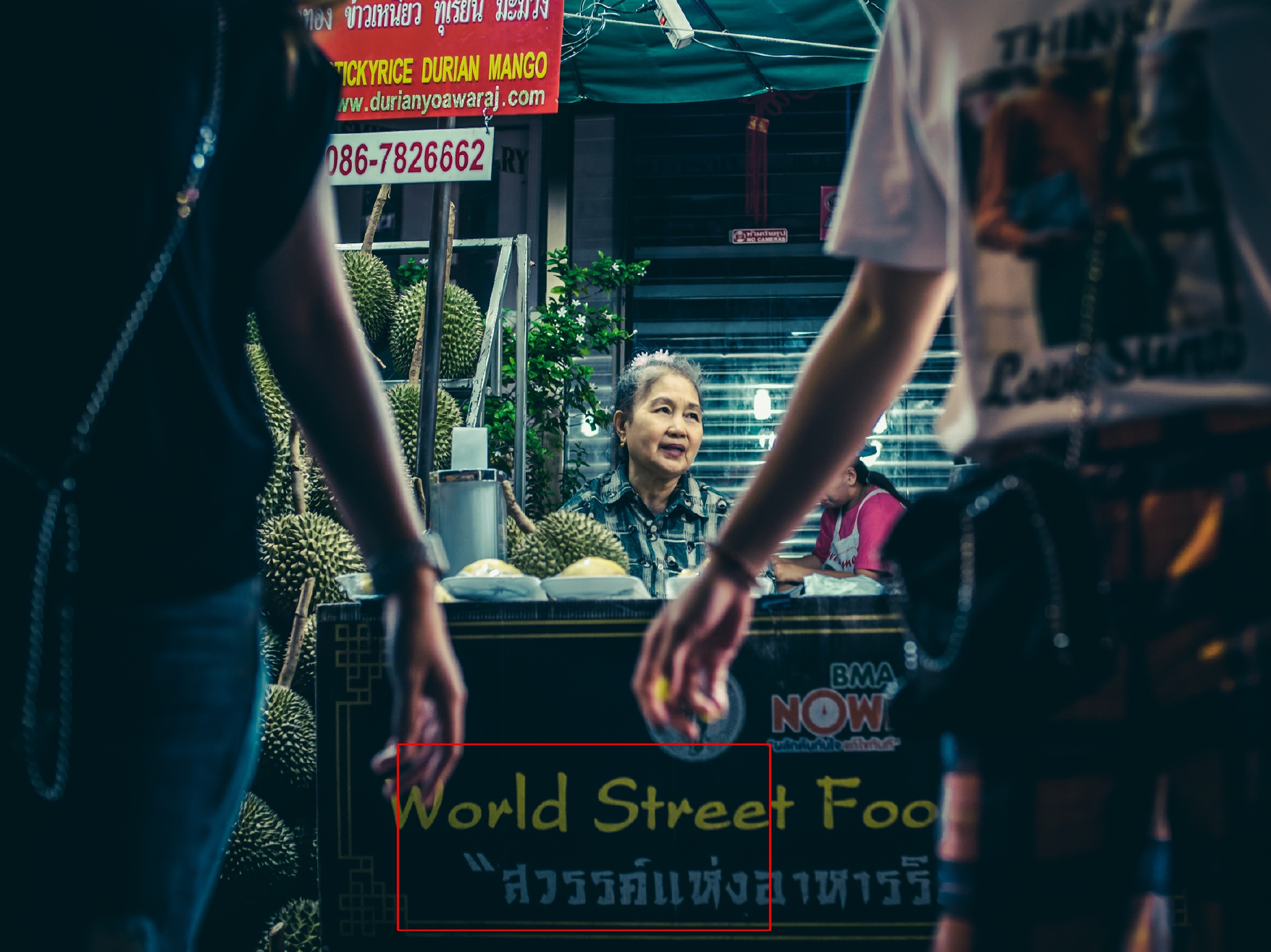}}}
            & \hspace{-4.0mm} \includegraphics[width=0.16\linewidth,height=0.105\linewidth]{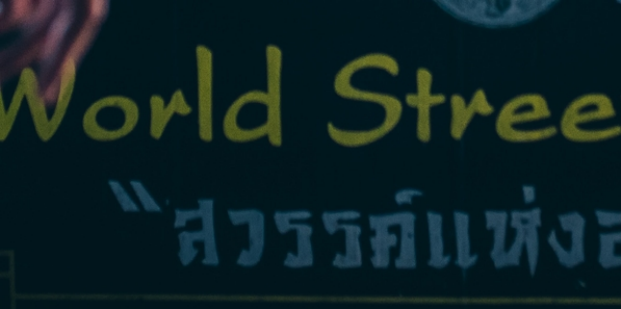}
            & \hspace{-4.0mm} \includegraphics[width=0.16\linewidth,height=0.105\linewidth]{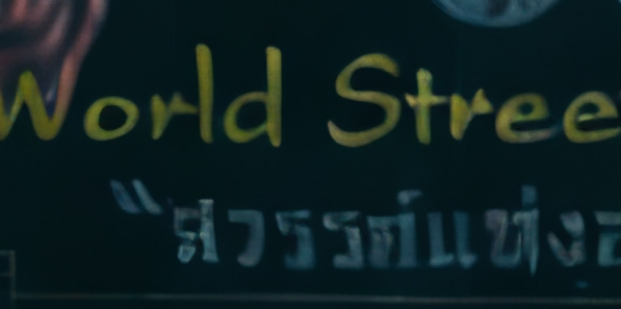}
            & \hspace{-4.0mm} \includegraphics[width=0.16\linewidth,height=0.105\linewidth]{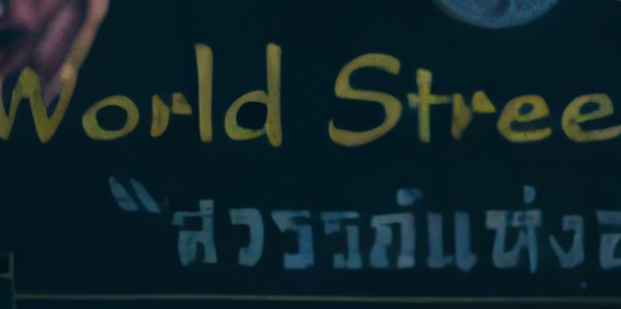}
            & \hspace{-4.0mm} \includegraphics[width=0.16\linewidth,height=0.105\linewidth]{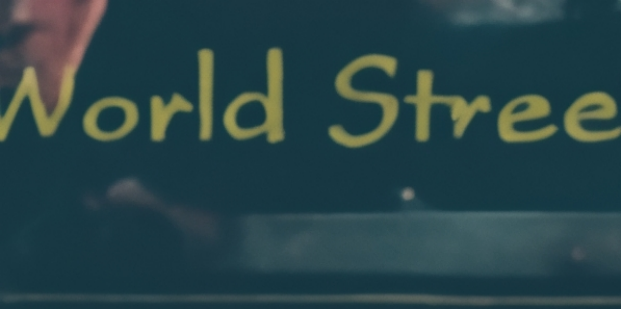}
              \\
    		\multicolumn{3}{c}{~}
            & \hspace{-4.0mm} (a) Original patch (24)
            & \hspace{-4.0mm} (b) Mao et al. (0.0273)
            & \hspace{-4.0mm} (c) GLC (0.0199)
            & \hspace{-4.0mm} (d) DiffEIC (0.0140) \\		
    	\multicolumn{3}{c}{~}
            & \hspace{-4.0mm} \includegraphics[width=0.16\linewidth,height=0.105\linewidth]{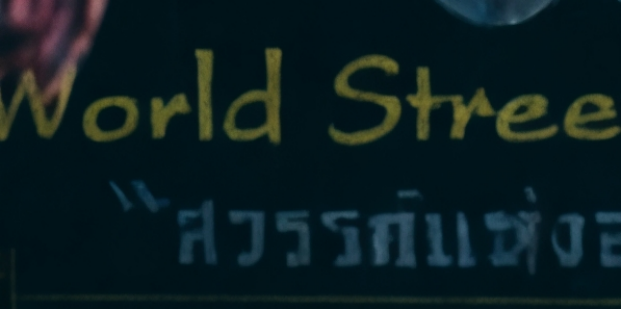}
            & \hspace{-4.0mm} \includegraphics[width=0.16\linewidth,height=0.105\linewidth]{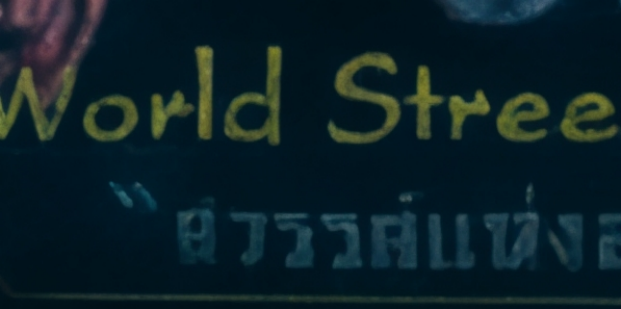}
            & \hspace{-4.0mm} \includegraphics[width=0.16\linewidth,height=0.105\linewidth]{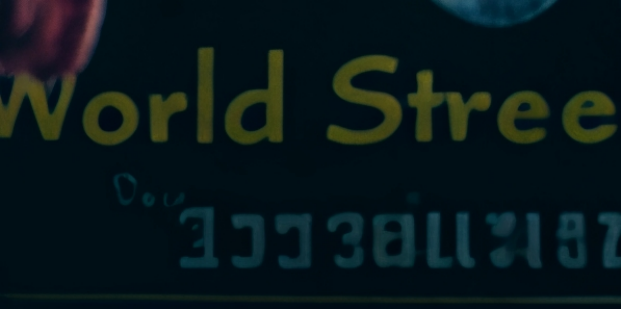}
            & \hspace{-4.0mm} \includegraphics[width=0.16\linewidth,height=0.105\linewidth]{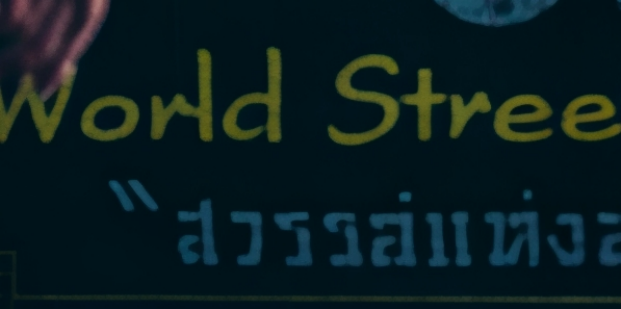}
            \\
    	\multicolumn{3}{c}{\hspace{-4.0mm} Original image}
            & \hspace{-4.0mm} (e) ResULIC (0.0224)
            & \hspace{-4.0mm} (f) RDEIC (0.0177)
            & \hspace{-4.0mm} (g) StableCodec (0.0179)
            & \hspace{-4.0mm} (h) SPRDiff (\textbf{0.0122})\\

            \multicolumn{3}{c}{\multirow{5}*[45.6pt]{
            \hspace{-2.5mm} \includegraphics[width=0.325\linewidth,height=0.23\linewidth]{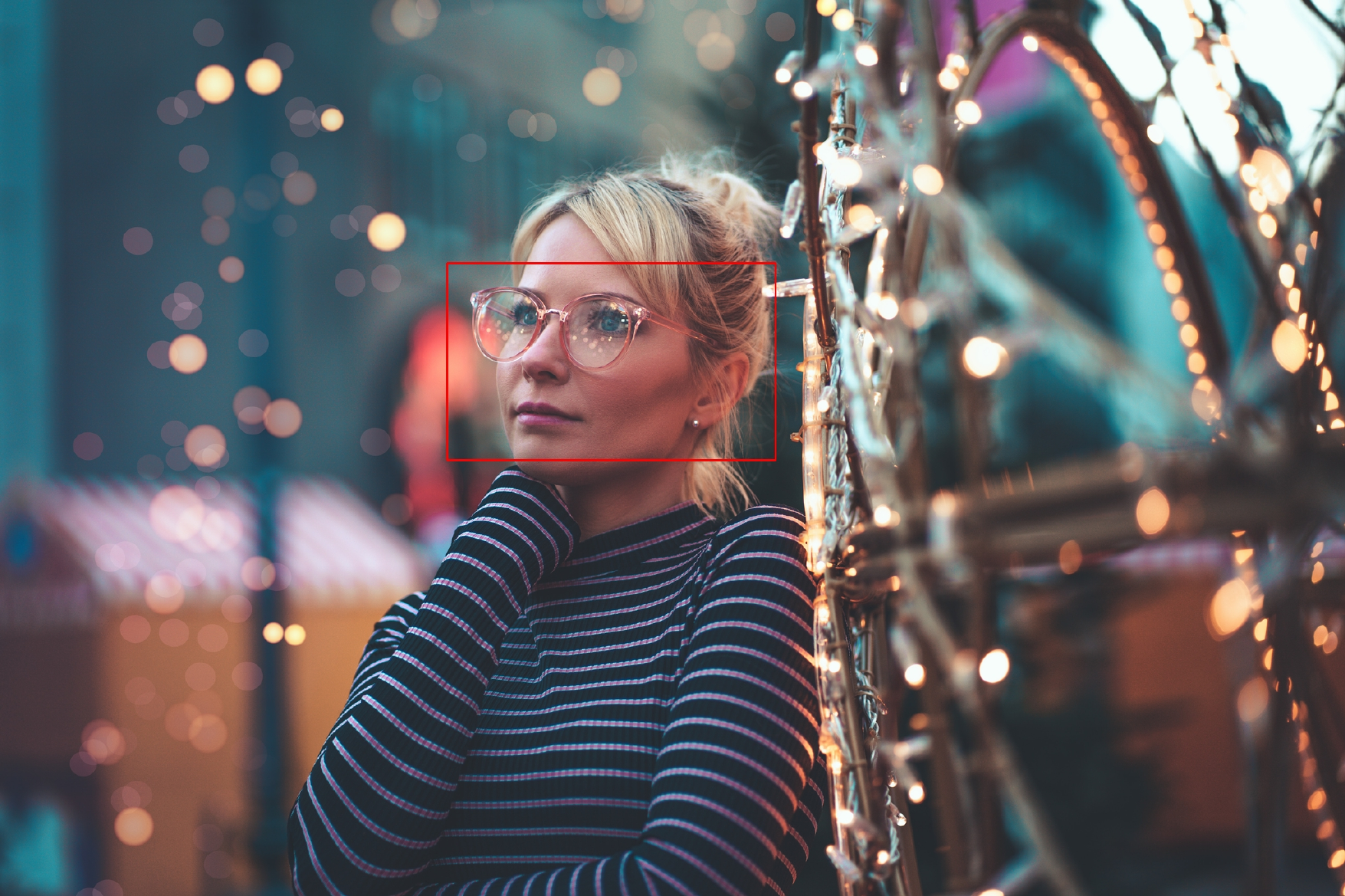}}}
            & \hspace{-4.0mm} \includegraphics[width=0.16\linewidth,height=0.105\linewidth]{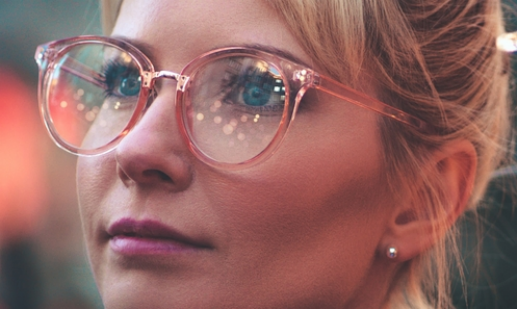}
            & \hspace{-4.0mm} \includegraphics[width=0.16\linewidth,height=0.105\linewidth]{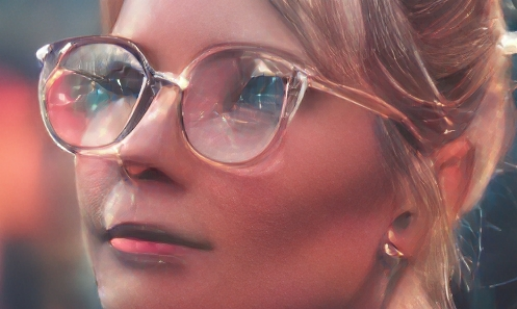}
            & \hspace{-4.0mm} \includegraphics[width=0.16\linewidth,height=0.105\linewidth]{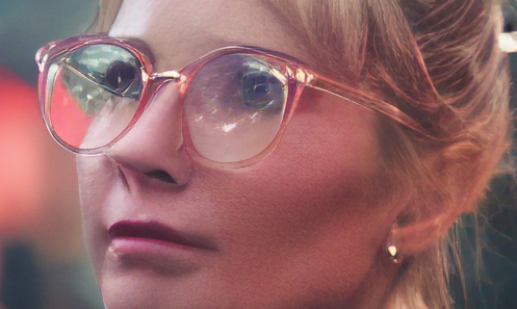}
            & \hspace{-4.0mm} \includegraphics[width=0.16\linewidth,height=0.105\linewidth]{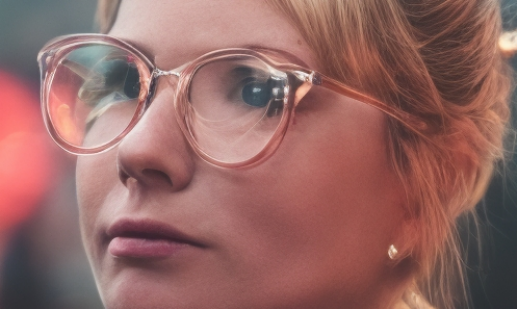}
              \\
    		\multicolumn{3}{c}{~}
            & \hspace{-4.0mm} (a) Original patch (24)
            & \hspace{-4.0mm} (b) Mao et al. (0.0201)
            & \hspace{-4.0mm} (c) GLC (0.0218)
            & \hspace{-4.0mm} (d) DiffEIC (0.0230) \\		
    	\multicolumn{3}{c}{~}
            & \hspace{-4.0mm} \includegraphics[width=0.16\linewidth,height=0.105\linewidth]{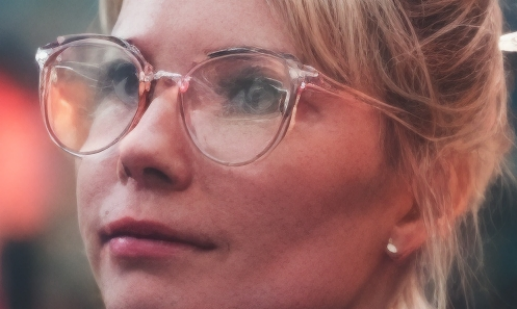}
            & \hspace{-4.0mm} \includegraphics[width=0.16\linewidth,height=0.105\linewidth]{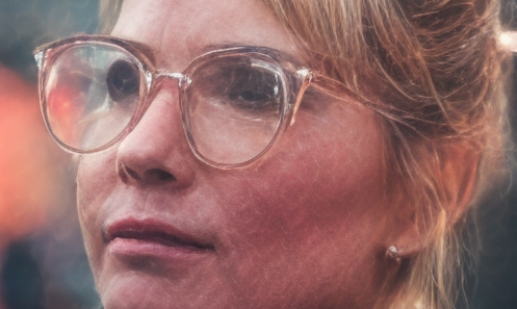}
            & \hspace{-4.0mm} \includegraphics[width=0.16\linewidth,height=0.105\linewidth]{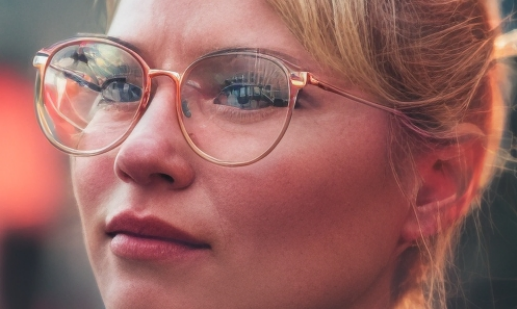}
            & \hspace{-4.0mm} \includegraphics[width=0.16\linewidth,height=0.105\linewidth]{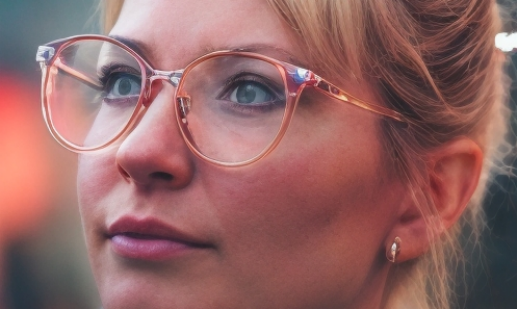}
            \\
    	\multicolumn{3}{c}{\hspace{-4.0mm} Original image}
            & \hspace{-4.0mm} (e) ResULIC (0.0231)
            & \hspace{-4.0mm} (f) RDEIC (0.0179)
            & \hspace{-4.0mm} (g) StableCodec (0.0214)
            & \hspace{-4.0mm} (h) SPRDiff (\textbf{0.0140}) \\

    \end{tabular}
\vspace{-2mm}
\caption{Visual comparisons with state-of-the-art methods on the CLIC2020 dataset. The value in parentheses denotes the bpp used for compression, with the best value marked in bold. Compared with other methods, the proposed SPRDiff produces higher-quality reconstructions with realistic details at lower bitrates.}
\label{vis_com}
\vspace{-3mm}
\end{figure*}
\subsubsection{Datasets}
Following~\cite{RDULIC_ICML2025}, we use the \emph{LSDIR}~\cite{LSDIR_CVPR2023} and \emph{Flicker2w}~\cite{Flicker2W} datasets, containing 105,736 images in total, for training. After training, we comprehensively evaluate the proposed method on three widely used benchmarks: the \emph{Kodak}~\cite{Kodak} dataset with 24 uncompressed natural images at a resolution of 768$\times$512, the \emph{CLIC2020}~\cite{CLIC2020} dataset with 428 high-quality images, and the \emph{Tecnick} dataset~\cite{Tecnick} with 100 images at a resolution of 1200$\times$1200.

\subsubsection{Evaluation Metrics} 
We adopt five metrics for quantitative evaluation, including perceptual fidelity metrics \emph{LPIPS}~\cite{LPIPS} and \emph{DISTS}~\cite{DISTS}, perceptual realism metrics \emph{FID}~\cite{FID} and \emph{KID}~\cite{KID}, and distortion metrics \emph{PSNR} and \emph{MS-SSIM}~\cite{MS-SSIM} to measure the quality of the reconstructed images. The FID and KID values are calculated using 256$\times$256 image patches, as described in~\cite{DiffEIC_TCSVT2024, DLF_Arxiv2025, ICISP_NN2025}. We do not report the FID and KID scores for the Kodak dataset as its small data size does not allow for a reliable evaluation. 

\subsubsection{Implementation Details}
\label{implemental_details}
We implement our method in PyTorch and train it from scratch on a single NVIDIA GeForce RTX 4090 GPU. During training, images are randomly cropped to 512$\times$512 resolution, and the batch size is set to 1. 
In the first training stage, we train the model for 30k iterations with a fixed learning rate of $e^{-4}$. The discriminator is disabled in this stage, and the hyperparameter $\lambda_b$ in Eq.~(\ref{stage1}) is 0.5. We adopt the pretrained SD-Turbo model~\cite{SD-Turbo_ECCV2024} as the generative prior and fine-tune it using LoRA~\cite{LoRA}. The LoRA rank is set to 16 for the VAE encoder and 32 for the denoising U-Net. It is worth noting that the weights of the distortion-oriented encoder and decoder are directly inherited from~\cite{MLIC++_2025}, while the semantic-oriented encoder corresponds to the pretrained DINOv2 model~\cite{DINOv2}.
In the second stage, we jointly fine-tune the compression network together with the discriminator for another 30k iterations. The learning rate is initialized at 5$\times e^{-5}$ and decays to 1$\times e^{-5}$ and 1$\times e^{-6}$ at the 28k and 29k iterations, respectively. The bitrate is controlled by setting $\lambda_b$ to $\{8, 4, 2\}$. The remaining hyperparameters are fixed across all experiments, with $\lambda_d=2$, $\lambda_{i}=2$, $\lambda_l=1$, and  $\lambda_a=0.1$.
\subsection{Comparisons with State-of-the-Art Methods}
We compare our method with state-of-the-art extreme image compression methods, including Mao et al.~\cite{Mao_DCC2024} and Murai~et~al.~\cite{Murai_VCIP2024}, GLC~\cite{GLC_CVPR2024}, ResULIC~\cite{RDULIC_ICML2025}, DiffEIC~\cite{DiffEIC_TCSVT2024}, RDEIC~\cite{RDEIC_TCSVT2025},  StableCodec~\cite{StableCodec_ICCV2025}, and OSCAR~\cite{OSCAR_Arxiv2025}.

\subsubsection{Quantitative Comparisons}
Fig.~\ref{quan_com} shows quantitative comparisons across the benchmark datasets. The proposed SPRDiff achieves the best perceptual and pixel-level fidelity on both the Kodak and Tecnick datasets. While RDEIC~\cite{RDEIC_TCSVT2025} achieves the highest PSNR on the CLIC2020 dataset, it sacrifices perceptual fidelity, with significantly high LPIPS and DISTS values compared to our SPRDiff. Furthermore, our method achieves LPIPS and DISTS values comparable to those of GLC~\cite{GLC_CVPR2024}, while outperforming it on PSNR and MS-SSIM. Additionally, StableCodec~\cite{StableCodec_ICCV2025} achieves competitive FID and KID values by leveraging the generative priors of pretrained SD-Turbo~\cite{SD-Turbo_ECCV2024}, but it underperforms in all other metrics. It is also worth noting that FID/KID are not particularly suitable for evaluating image compression performance, as they focus on the difference in distribution between reconstructed and original images. However, the compression task focuses on image-to-image reconstruction fidelity. All comparisons demonstrate that the proposed SPRDiff achieves a better distortion-perception trade-off at ultra-low bitrates. 

\subsubsection{Qualitative Comparisons}
Fig.~\ref{vis_com} shows visual comparisons from the CLIC2020 dataset. The token-based compression method proposed by Mao~et~al. produces severe artifacts at a low bitrate of around 0.02 bpp. Similarly, GLC struggles to generate realistic details, resulting in visible artifacts. The diffusion-based codecs DiffEIC, ResULIC, RDEIC, and StableCodec produce reconstructions that are either extremely blurry or fail to match the original images in terms of reconstruction fidelity. For instance, the texture of the cat's face often appears smooth, whereas the structure of the letters and face becomes degraded and distorted. By contrast, the proposed SPRDiff yields reconstructed results with high fidelity and realism, preserving texture and structural details at ultra-low bitrates.
\subsubsection{Model Complexity Comparison}
\begin{table*}[tbp]
\centering
\caption{Model complexity comparison. Inference time is measured on 512$\times$512 images using a machine with a single NVIDIA GeForce RTX 3090 GPU. Best results are in \textbf{bold}.}
\label{mcc} 
\begin{tabular}{l|c|ccc|ccccc}
\toprule
    & & \multicolumn{3}{c|}{Inference time (ms)} & & & & &  \\
    \cline{3-5}
    Methods & Parameters (M) & Encoding & Decoding & Average & Bpp & PSNR &  MS-SSIM & LPIPS & DISTS \\
    \midrule
    DiffEIC~\cite{DiffEIC_TCSVT2024}  & 1,379.50 & 157.80 & 3,734.19 & 1,946.00 & 0.0378 & 20.71 & 0.7526 & 0.3517 & 0.1386 \\
    RDEIC ~\cite{RDEIC_TCSVT2025}   & 1,380.27 & 135.09 & 257.51 & 193.60 & \textbf{0.0249} & 22.44 & 0.7662 & 0.3635 & 0.1613 \\ 
    OSCAR~\cite{OSCAR_Arxiv2025}   & \textbf{1,009.30} & \textbf{107.81} & 287.49 & 197.65 & 0.0313 & 20.77 & 0.7439  & 0.3609 & 0.1387 \\
    StableCodec~\cite{StableCodec_ICCV2025} &  1,065.81 & 115.68 & \textbf{215.29} & \textbf{165.49} & 0.0303 & 22.11 &  0.7662 & 0.3128 & 0.1158 \\
    \midrule
    SPRDiff (Ours) & 1,344.23 & 200.87 & 330.92 & 265.90 & 0.0285 & \textbf{23.15} & \textbf{0.8133} & \textbf{0.2793} & \textbf{0.1072} \\
    \bottomrule
\end{tabular}
\end{table*}
We further compare the model complexity in Table~\ref{mcc}. All experiments are conducted on 512$\times$512 images using a single RTX~3090 GPU. Our SPRDiff contains fewer parameters and achieves a faster inference speed than DiffEIC~\cite{DiffEIC_TCSVT2024}, while exhibiting higher complexity compared to StableCodec~\cite{StableCodec_ICCV2025}.
\subsection{Analysis and Discussion}
\subsubsection{Effect of Triple-encoder}
\begin{figure}[htbp]
\centering
\includegraphics[width=0.45\textwidth]{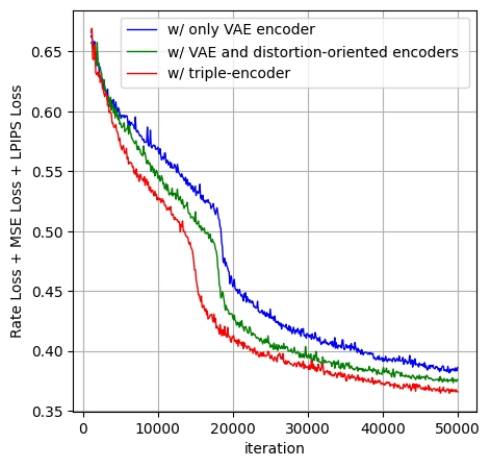}\vspace{-0.54cm}
\caption{The rate–distortion–perception validation loss curve for the first training stage is presented. Using the proposed triple-encoder facilitates convergence to a lower overall rate–distortion–perception loss.} 
\label{enc_loss}
\end{figure}

To further validate the effectiveness of the proposed triple-encoder, we compare it against two baseline configurations: one that employs only the pretrained VAE encoder, and another that combines the VAE and distortion-oriented encoders. Fig.~\ref{enc_loss} shows that the proposed triple-encoder achieves convergence to a lower rate–distortion–perception loss compared with the baseline configurations. Table~\ref{te_com} further demonstrates that the proposed triple-encoder yields notable performance gains, achieving superior perceptual and pixel-wise fidelity metrics compared with the baseline methods.
\begin{table}[htbp]
\centering
\caption{Ablation on the triple-encoder. The images are compressed at 0.02 bpp. Best results are highlighted in \textbf{bold}.}
\label{te_com} 
\begin{tabular}{l|ccccc}
\toprule
    Methods & PSNR $\uparrow$ & MS-SSIM $\uparrow$ & LPIPS $\downarrow$ & DISTS $\downarrow$  \\
    \midrule
    w/ only $\mathcal{E}_{vae}$ & 21.89 & 0.7590 & 0.3213 & 0.1148 \\ 
    w/ $\mathcal{E}_{vae}$ \& $\mathcal{E}_{doe}$ & 22.01 & 0.7637 & 0.3173 & 0.1135 \\ 
    \midrule
    w/ triple-encoder & \textbf{22.12} & \textbf{0.7687} & \textbf{0.3106} & \textbf{0.1106} \\ 
    \bottomrule
\end{tabular}
\end{table}
\subsubsection{Effectiveness of Semantic Representation}
\begin{figure}[htbp]
\centering
\includegraphics[width=0.48\textwidth]{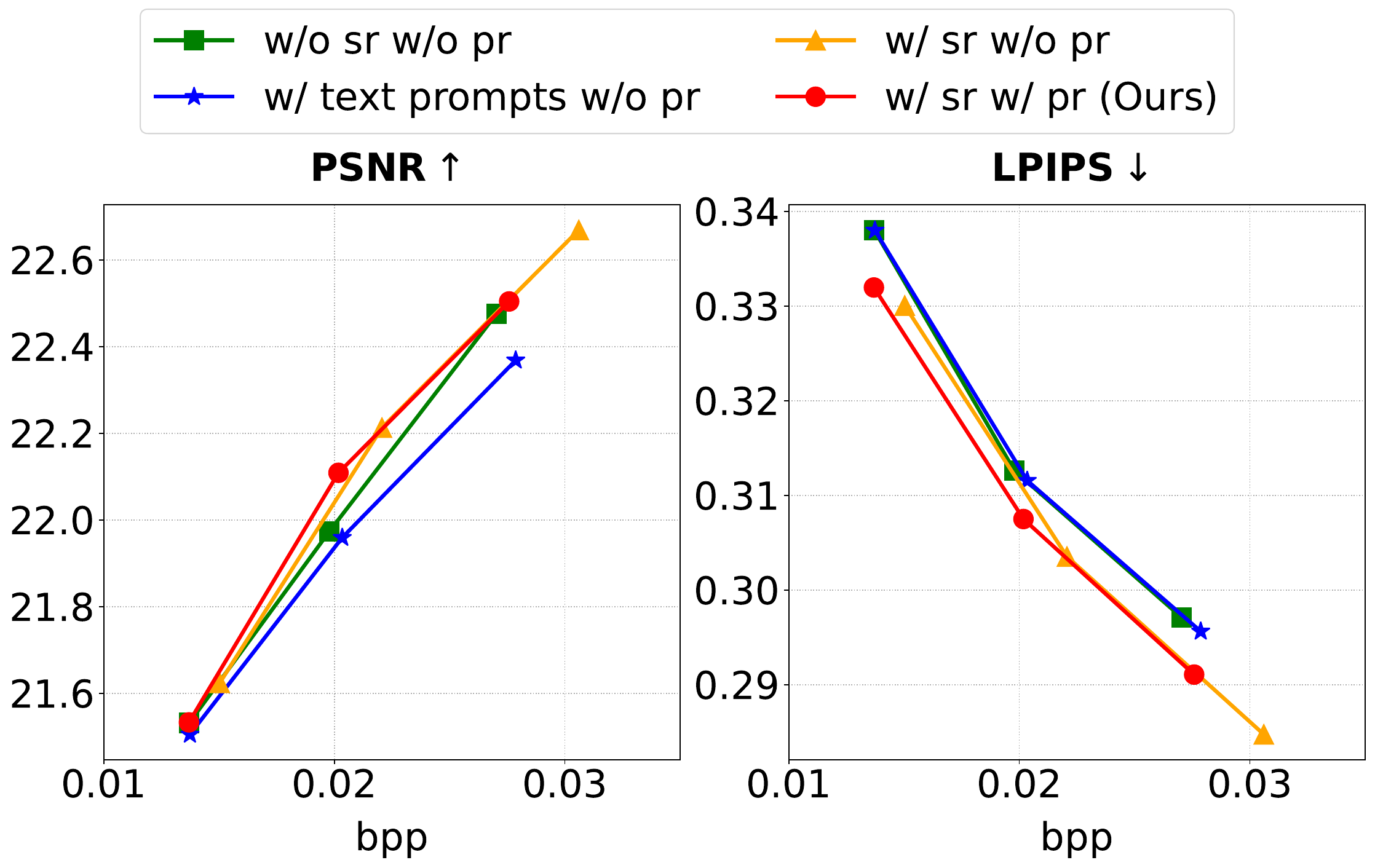}\vspace{-0.2cm}
\caption{Ablation studies on the semantic and pixel-level representation (denoted as sr and pr, respectively) for extreme image compression. Higher PSNR and lower LPIPS indicate better reconstruction performance.} 
\label{ab_com}
\end{figure}
\begin{figure}[!t]
\footnotesize
\centering
    \begin{tabular}{c c c c}
            \includegraphics[width=0.24\linewidth,height=0.15\linewidth]{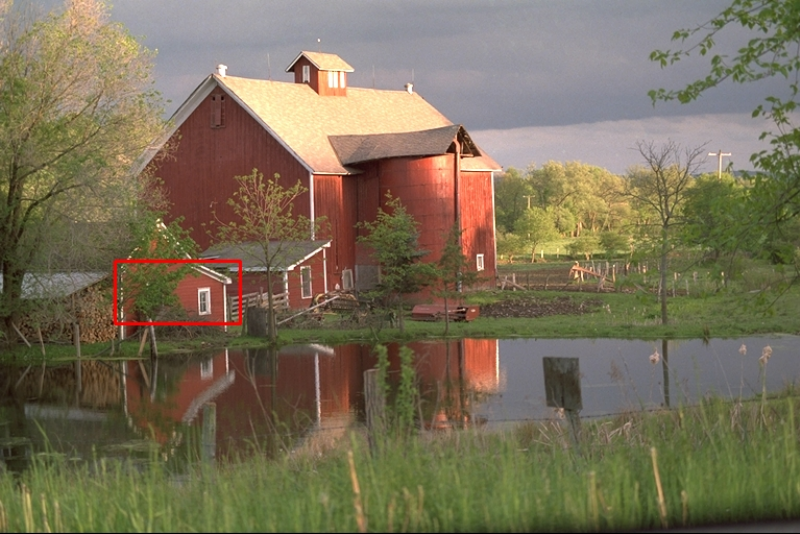}
            \includegraphics[width=0.24\linewidth,height=0.15\linewidth]{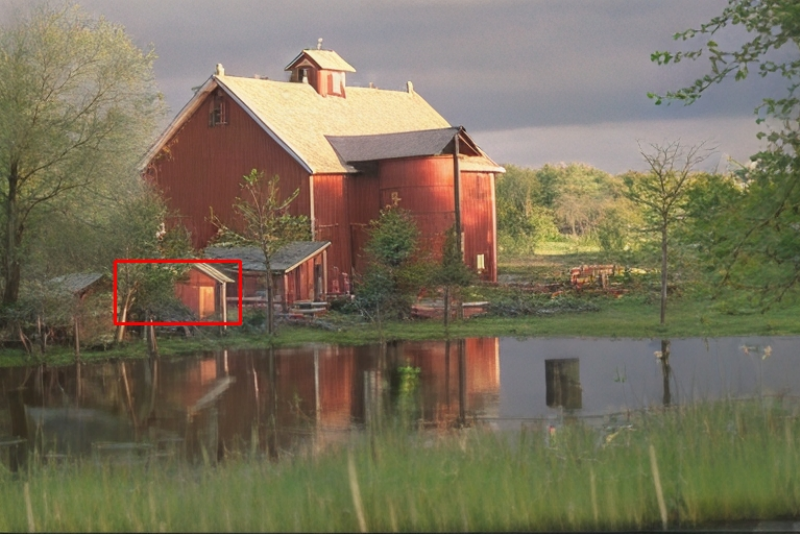}
            \includegraphics[width=0.24\linewidth,height=0.15\linewidth]{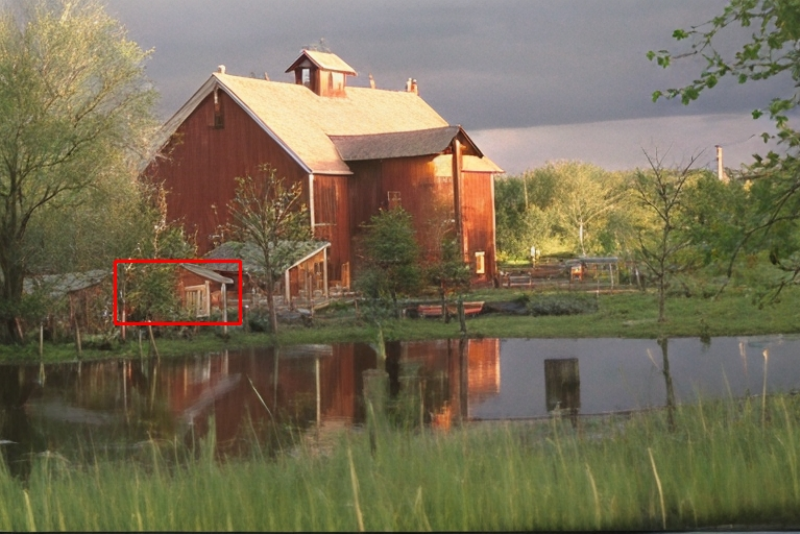}
            \includegraphics[width=0.24\linewidth,height=0.15\linewidth]{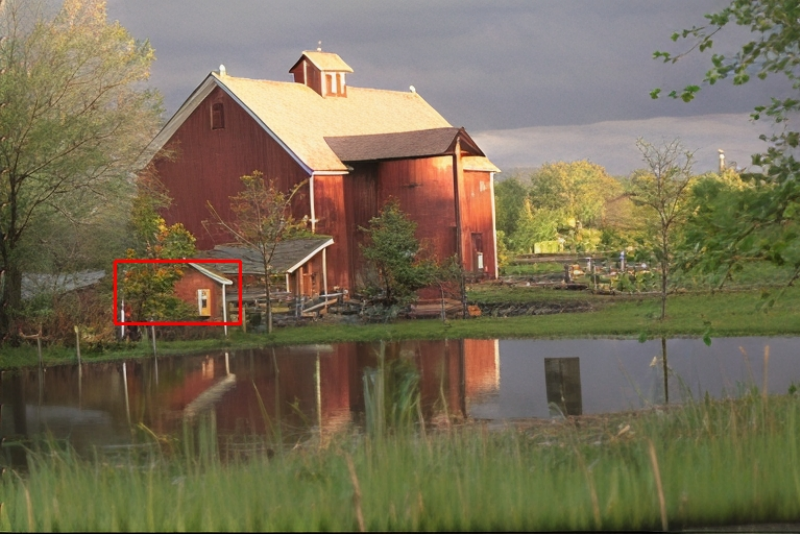} \\
            \includegraphics[width=0.24\linewidth,height=0.15\linewidth]{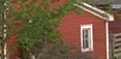}
            \includegraphics[width=0.24\linewidth,height=0.15\linewidth]{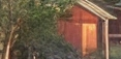}
            \includegraphics[width=0.24\linewidth,height=0.15\linewidth]{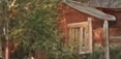}
            \includegraphics[width=0.24\linewidth,height=0.15\linewidth]{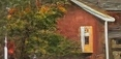} \\
            \makebox[0.24\linewidth]{(a) Original}
            \makebox[0.24\linewidth]{(b) w/o $F_s$}
            \makebox[0.24\linewidth]{(c) w/ text prompts}
            \makebox[0.24\linewidth]{(d) w/ $F_s$}	 \\
    \end{tabular}
\vspace{-2mm}
\caption{Effectiveness of semantic representation $F_s$. (b) shows the reconstruction obtained without any semantic conditioning. (c) and (d) presents reconstructions conditioned on text prompts and on the proposed $F_s$, respectively. All results are generated under a bitrate of 0.02 bpp.}
\label{sp_com}
\vspace{-3mm}
\end{figure}
Different from existing extreme compression approaches that employ text prompts as semantic guidance for diffusion models~\cite{PerCo_ICLR2023, StableCodec_ICCV2025}, we instead exploit semantic features $F_s$ extracted from the intermediate reconstructed images using a pretrained semantic-oriented encoder, as formulated in Eq.(\ref{dor}). To evaluate whether $F_s$ facilitates extreme image compression, we construct two variants of the proposed SPRDiff without $F_s$. The first variant excludes all semantic representations, while the second leverages a CLIP model~\cite{CLIP} to extract semantic features from a robust text prompt (e.g., ``a high-resolution, 8K, ultra-realistic image with sharp focus, vibrant colors, and natural lighting").

As shown in Fig.~\ref{ab_com}, using text prompts as in~\cite{StableCodec_ICCV2025} fails to improve perceptual fidelity and even degrades pixel-level accuracy, as evidenced by its lower PSNR compared to the baseline without semantic representations (green curve vs. blue curve). In contrast, incorporating the proposed semantic representation yields consistently better reconstructions at ultra-low bitrates, achieving higher PSNR and lower LPIPS values (orange curve).

In addition, Fig.~\ref{sp_com}(b) and (c) illustrate that both the method without semantic conditioning and the one conditioned on text prompts exhibit noticeable semantic drift. By contrast, the proposed approach leveraging $F_s$ produces substantially clearer reconstructions, where the structural boundaries of the house are faithfully preserved (see Fig.~\ref{sp_com}(d)).
\subsubsection{Impact of Pixel Representation}
\begin{figure}[!t]
\footnotesize
\centering
    \begin{tabular}{c c c}
            \includegraphics[width=0.32\linewidth,height=0.4\linewidth]{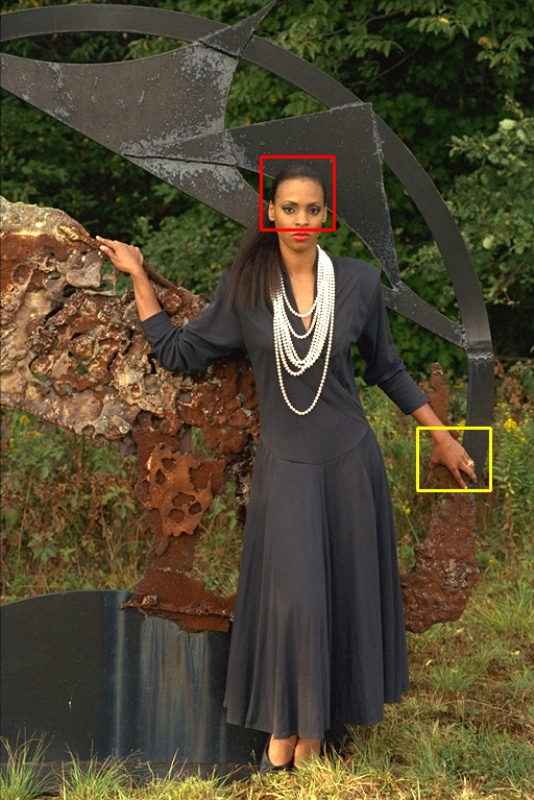}
            \includegraphics[width=0.32\linewidth,height=0.4\linewidth]{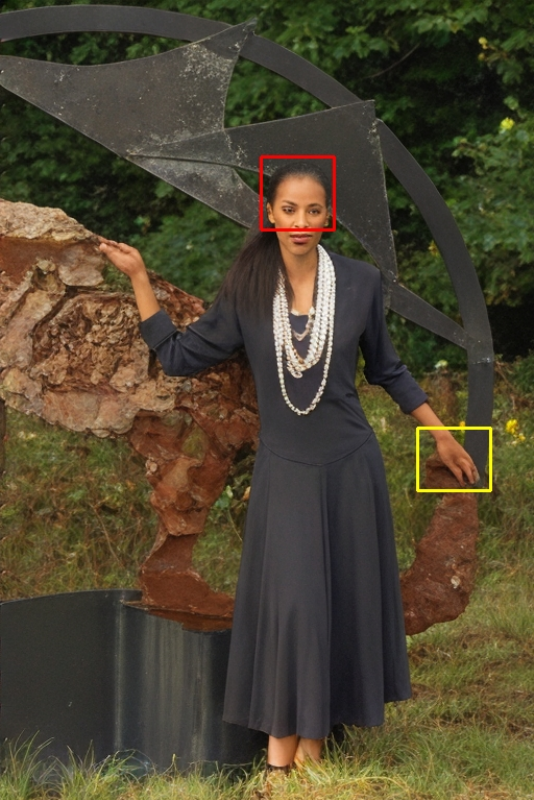}
            \includegraphics[width=0.32\linewidth,height=0.4\linewidth]{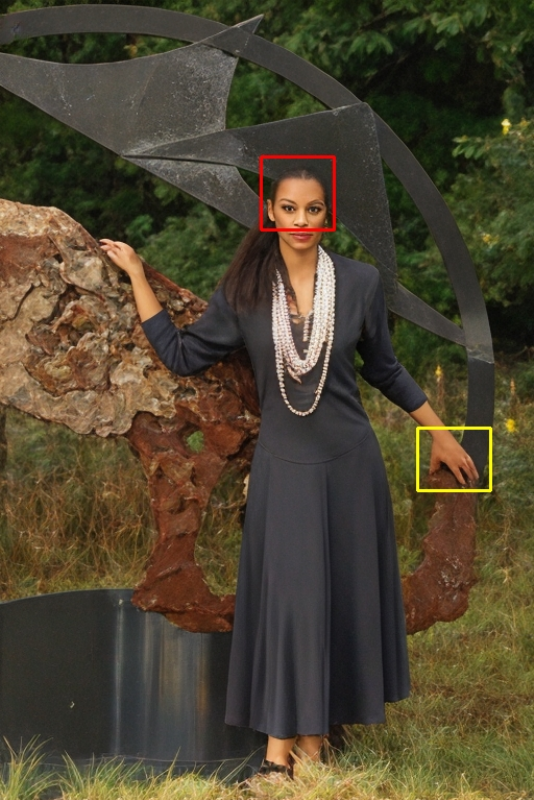} \\
            \includegraphics[width=0.19\linewidth,height=0.15\linewidth]{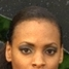}
            \includegraphics[width=0.12\linewidth,height=0.15\linewidth]{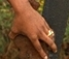}
             \includegraphics[width=0.19\linewidth,height=0.15\linewidth]{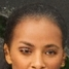}
            \includegraphics[width=0.12\linewidth,height=0.15\linewidth]{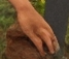}
            \includegraphics[width=0.19\linewidth,height=0.15\linewidth]{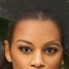}
            \includegraphics[width=0.12\linewidth,height=0.15\linewidth]{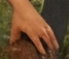} \\
            \makebox[0.32\linewidth]{(a) Original}
            \makebox[0.32\linewidth]{(b) w/o $F_p$}
            \makebox[0.32\linewidth]{(c) w/ $F_p$}	 \\
    \end{tabular}
\vspace{-2mm}
\caption{Impact of pixel representation $F_p$. (b) and (c) are reconstructions using the method without and with pixel guidance at around 0.02 bpp, respectively.}
\label{pg_com}
\vspace{-3mm}
\end{figure}
\begin{figure}[!t]
\footnotesize
\centering
    \begin{tabular}{cccccc}
            \multicolumn{6}{c}{
            \includegraphics[width=0.98\linewidth,height=0.6\linewidth]{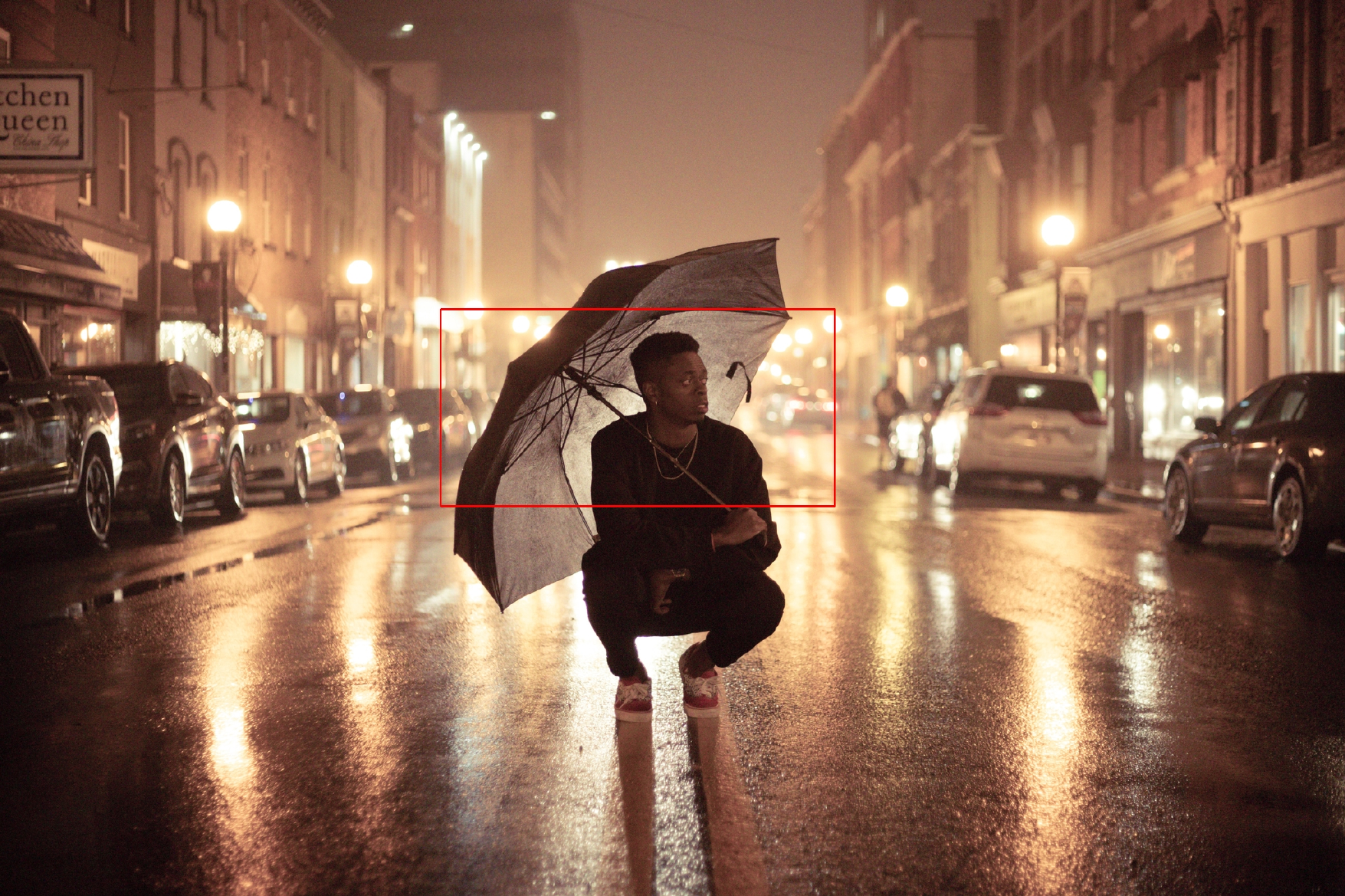}} \\
            \makebox[0.98\linewidth]{Original image} \\
            \includegraphics[width=0.32\linewidth,height=0.21\linewidth]{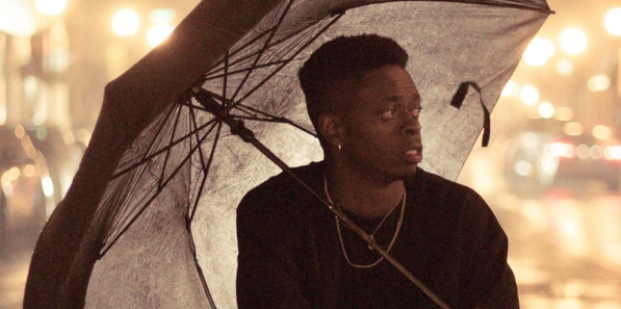}
            \includegraphics[width=0.32\linewidth,height=0.21\linewidth]{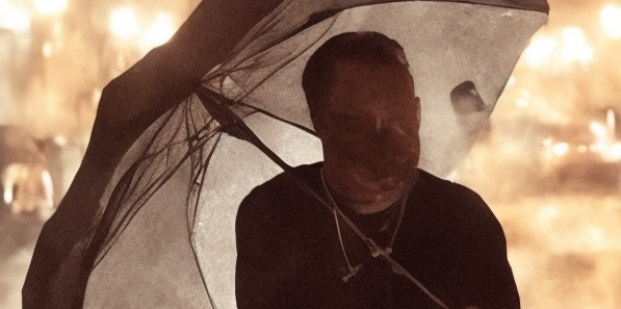}
            \includegraphics[width=0.32\linewidth,height=0.21\linewidth]{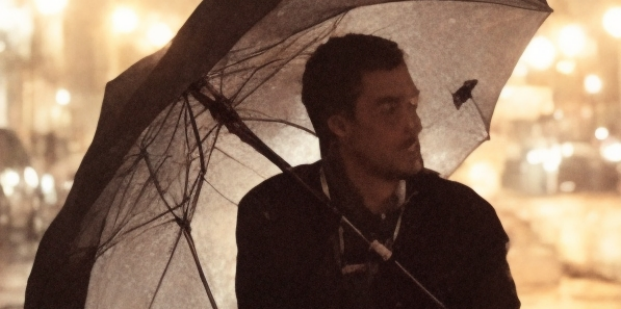}
              \\
            \makebox[0.32\linewidth]{(a) Original patch (24)}
            \makebox[0.32\linewidth]{(b) DiffEIC (0.0145)} 
            \makebox[0.32\linewidth]{(c) ResULIC (0.0125)} \\
            \includegraphics[width=0.32\linewidth,height=0.21\linewidth]{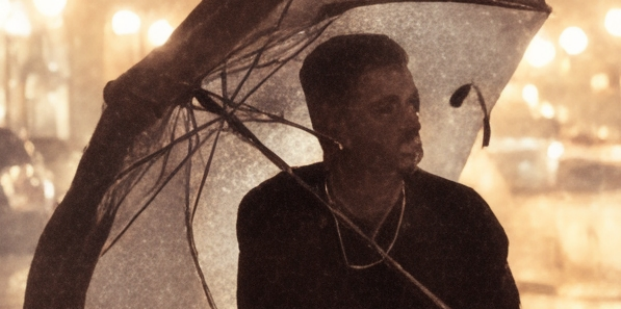}
            \includegraphics[width=0.32\linewidth,height=0.21\linewidth]{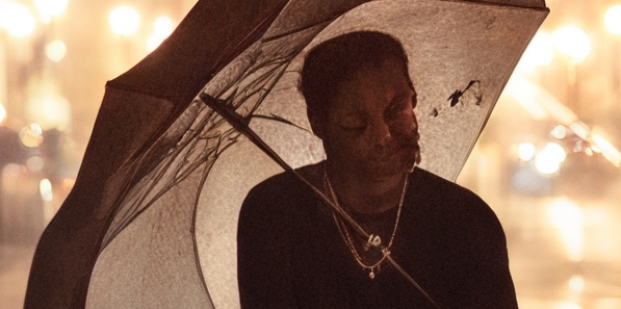}
            \includegraphics[width=0.32\linewidth,height=0.21\linewidth]{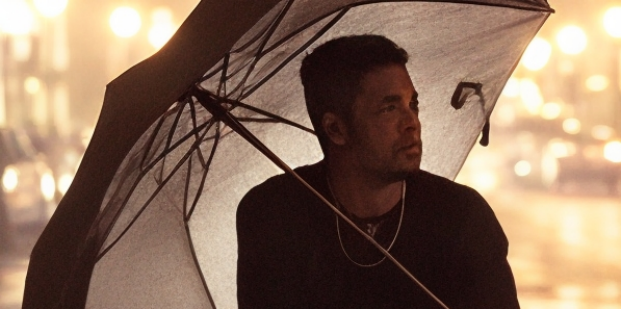} \\
            \makebox[0.32\linewidth]{(d) RDEIC (0.0166)}
            \makebox[0.32\linewidth]{(e) StableCodec (0.0189)}
            \makebox[0.32\linewidth]{(f) SPRDiff (\textbf{0.0116})}\\
    \end{tabular}
\vspace{-2mm}
\caption{Limitation: Facial reconstruction in complex backgrounds. The value in parentheses denotes the bpp; the best value is marked in bold. Although the proposed SPRDiff can generate visually pleasing results at lower bitrates than other methods, the face identity is altered and deviates from the original.}
\label{limitation}
\vspace{-3mm}
\end{figure}
The proposed SPRDiff leverages the pixel-level representation $F_p$, derived from Eq.(\ref{dor}), to refine the noisy latent in the diffusion model and thereby enhance the fidelity of pixel-wise reconstruction. To assess the contribution $F_p$, we remove the lightweight projector $\mathcal{P}$ and retrain the model under identical experimental configurations for a fair comparison. The quantitative results are presented in Fig.~\ref{ab_com}. 
Using $F_p$ yields consistent improvements in compression performance, particularly at low bitrates (red curve vs. orange curve). Furthermore, as illustrated in Fig.~\ref{pg_com}, the adoption of the pixel-level representation leads to improved reconstruction fidelity, enabling a more faithful reconstruction of fine structural details, such as the eyes and hands.
\subsubsection{Limitations}
Fig.~\ref{limitation} shows visual comparisons of face reconstruction at extremely low bitrates. In this image, the face occupies only a small part of the image, while the background is complex. We find that the proposed SPRDiff outperforms competing extreme compression methods. However, the face's identity is modified, deviating from the original (see Fig.~\ref{limitation}(a) vs. (f)). Future work will investigate an ROI-based extreme compression model to address this limitation. Moreover, given the model complexity of the proposed SPRDiff, exploring more lightweight solutions for ultra-low bitrate compression represents another promising direction, for instance, incorporating model compression techniques such as pruning and distillation.

\section{Conclusion}
\label{conclusion}
In this paper, we propose an effective extreme image compression framework, termed SPRDiff, which leverages both semantic and pixel-level representations to enhance the reconstruction fidelity at ultra-low bitrates. 
We first introduce a triple-encoder architecture that compensates for the substantial information loss inherent in the pretrained VAE encoder by integrating distortion-oriented and semantic-oriented encoders. 
We further design a distortion-aware reconstruction module with dual-feature extraction, which delivers informative semantic and pixel conditional cues to guide the diffusion model toward high-fidelity reconstruction. 
Extensive quantitative and qualitative results demonstrate that the proposed method achieves a superior rate-distortion-perception tradeoff compared to state-of-the-art approaches.

\bibliographystyle{IEEEtran}
\bibliography{main}

\begin{IEEEbiography}[{\includegraphics[width=1in,height=1.25in,clip,keepaspectratio]{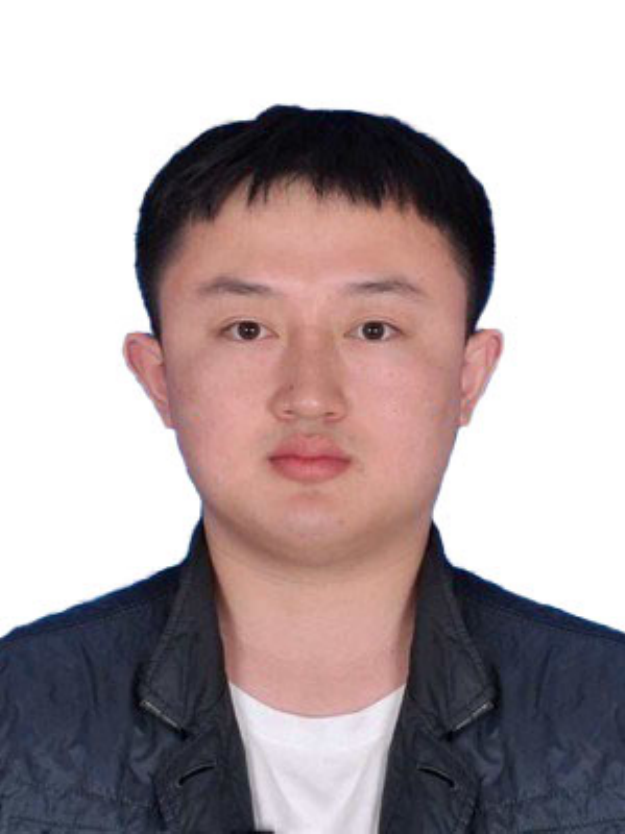}}]{Hao Wei}
is currently a Ph.D. candidate with the Institute of Artificial Intelligence and Robotics at Xi'an Jiaotong University, and a visiting Ph.D. student at the University of Western Australia, supervised by Professor Ajmal Mian. He received his B.Sc. and M.Sc. degrees from Yangzhou University and Nanjing University of Science and Technology in 2018 and 2021, respectively. His research interests include image deblurring, compression, rescaling, and other low-level vision problems. He is also a reviewer for TCSVT, TMM, CVPR, ICCV, ICLR and many other journals and conferences.
\end{IEEEbiography}

\begin{IEEEbiography}
[{\includegraphics[width=1in,height=1.25in,clip,keepaspectratio]{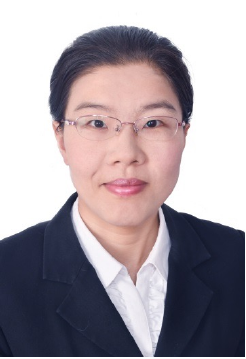}}]{Yanhui Zhou}
received the M.S. and Ph.D. degrees in electrical engineering from the Xi'an Jiaotong University, Xi'an, China, in 2005 and 2011, respectively. She is currently an associate professor with the School of Information and Telecommunication at Xi’an Jiaotong University. Her current research interests include image/video compression, computer vision and deep learning.
\end{IEEEbiography}

\begin{IEEEbiography}[{\includegraphics[width=1in,height=1.25in,clip,keepaspectratio]{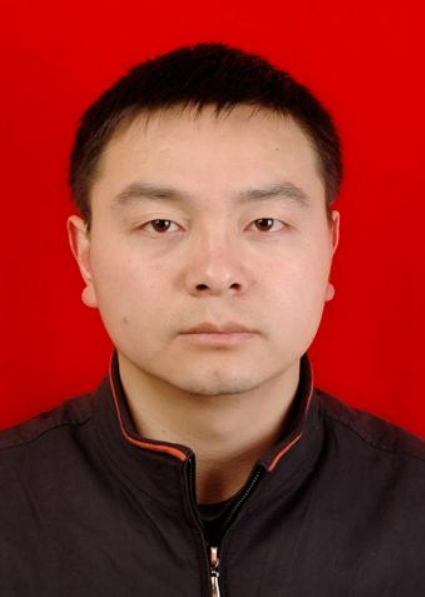}}]{Chenyang Ge}
is currently a professor at Xi'an Jiaotong University. He received the B.A., M.S., and Ph.D. degrees at Xi'an Jiaotong University in 1999, 2002, and 2009, respectively. His research interests include computer vision, 3D sensing, new display processing, and SoC design.
\end{IEEEbiography}

\begin{IEEEbiography}
[{\includegraphics[width=1in,height=1.25in,clip,keepaspectratio]{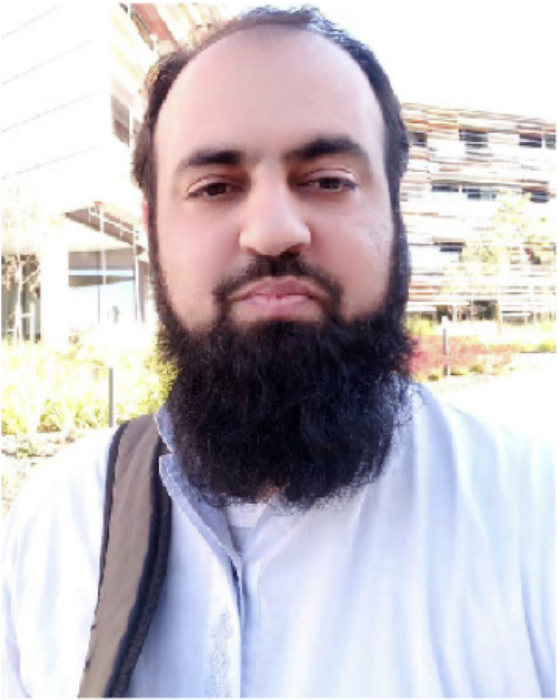}}]{Saeed Anwar} is a Senior Lecturer of Computer Vision, Machine Learning, AI, and Data Science at the Department of Computer Science and Software Engineering, University of Western Australia. He also holds honorary affiliations at the Australian National University and CSIRO. Dr. Anwar has received best paper awards at ICPR'24 and RS'22, and has been nominated for best paper awards at CVPR'20 and PR'19. He currently serves as an associate editor at the IEEE Journal of Oceanic Engineering (JOE) and has previously served as an associate editor for Neurocomputing. His editorial contributions includes guest editorials for the IEEE's JOE, MDPI's Sensors, and Frontiers in Robotics and AI. He has served as an area chair for CVPR'26, WACV'26, ICCV'25, IJCAI'25, and ACCV'24. He regularly reviews submissions for top venues in his field, such as TPAMI, TIP, CVPR, ICCV, ECCV, ACM MM, and many others. He has worked on multiple commercial projects as a team lead with Boeing, LionShare, and Bionic Vision. He has also supervised PhD and Master's students through to completion.
\end{IEEEbiography}

\begin{IEEEbiography}[{\includegraphics[width=1in,height=1.25in,clip,keepaspectratio]{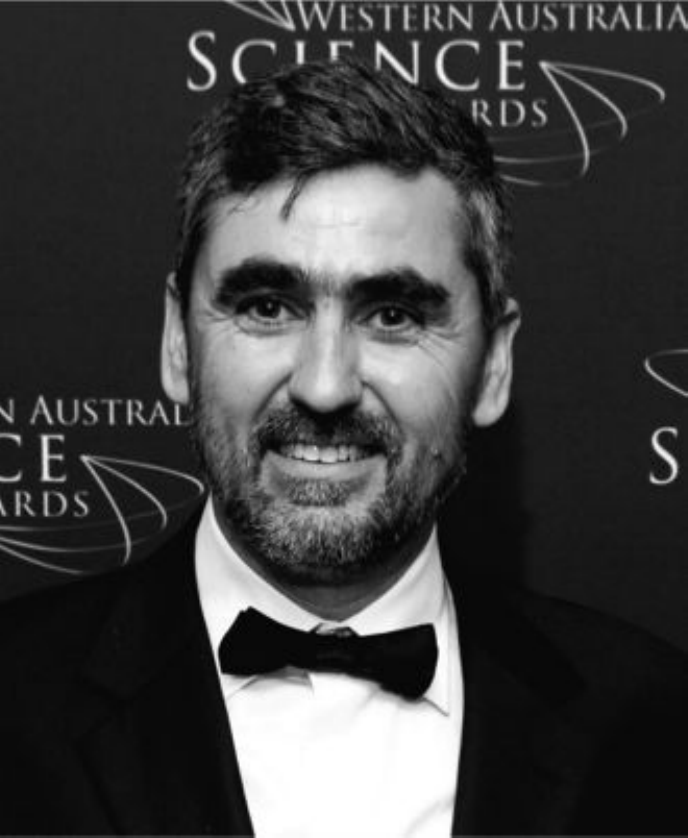}}]{Ajmal Mian}(Senior Member, IEEE)
is a Professor of Computer Science at The University of Western Australia. He is the recipient of three esteemed fellowships from the Australian Research Council (ARC). He has also received several research grants from the ARC, the National Health and Medical Research Council of Australia, US Department of Defense and the Australian Department of Defense with a combined funding of over \$41 million. He received the West Australian Early Career Scientist of the Year 2012 award and the HBF Mid-career Scientist of the Year 2022 award. He has also received several other awards including the Excellence in Research Supervision Award, EH Thompson Award, ASPIRE Professional Development Award, Vice-chancellors Mid- career Award, Outstanding Young Investigator Award, and the Australasian Distinguished Dissertation Award. He is an IAPR Fellow and Distinguished Speaker of the ACM. He is an Associate Editor of IEEE Transactions on Pattern Analysis and Machine Intelligence and previously served as a Senior Editor for IEEE Transactions in Neural Networks and Learning Systems and Associate Editor for IEEE Transactions on Image Processing and the Pattern Recognition journal. He was the General Co-Chair of DICTA 2024, DICTA 2019 and ACCV 2018. His research interests are in 3D computer vision, machine learning, and video analysis.
\end{IEEEbiography}

%
\IEEEpeerreviewmaketitle

\ifCLASSOPTIONcaptionsoff
  \newpage
\fi

\end{document}